\documentclass{article}
\usepackage{iclr2027_conference,times}

\usepackage[utf8]{inputenc}
\usepackage[T1]{fontenc}
\usepackage[table]{xcolor}
\newcommand{\revision}[1]{#1}
\newcommand{\currentrevision}[1]{#1}
\usepackage{float}
\usepackage{wrapfig}
\makeatletter   
\newcommand{\needspace}[1]{\par\penalty-100\begingroup\setlength{\dimen@}{#1}\dimen@ii\pagegoal\advance\dimen@ii-\pagetotal\ifdim\dimen@>\dimen@ii\ifdim\dimen@ii>\z@\vfil\fi\break\fi\endgroup}
\makeatother
\usepackage{placeins}
\usepackage{url}
\usepackage{booktabs}
\usepackage{multirow}
\usepackage{graphicx}
\usepackage{amsmath,amssymb}
\usepackage{pifont}
\usepackage{etoolbox}
\usepackage{enumitem}
\usepackage[hypertexnames=false]{hyperref}   
\hypersetup{colorlinks=true,linkcolor=[HTML]{1F4E9A},citecolor=[HTML]{1F4E9A},urlcolor=[HTML]{1F4E9A}}

\graphicspath{{figures/}}
\newcommand{\cmark}{\ding{51}}

\definecolor{rankA}{HTML}{F7B2B2}\definecolor{rankB}{HTML}{FBD3A6}\definecolor{rankC}{HTML}{FFF1AE}
\newcommand{\bestc}[1]{\cellcolor{rankA}\textbf{#1}}
\newcommand{\secondc}[1]{\cellcolor{rankB}#1}
\newcommand{\thirdc}[1]{\cellcolor{rankC}#1}
\newcommand{\ranklegendF}{{\setlength{\fboxsep}{1.2pt}\scriptsize\colorbox{rankA}{\strut\textbf{best}}\,\colorbox{rankB}{\strut 2nd}\,\colorbox{rankC}{\strut 3rd}}}

\AtBeginEnvironment{table}{\setlength{\abovecaptionskip}{0pt}\setlength{\belowcaptionskip}{4pt}}
\DeclareRobustCommand{\name}{\textit{WildHSR}}

\newcommand{\oursWA}{66.3}
\newcommand{\oursW}{193.6}
\newcommand{\oursRTE}{0.90}
\newcommand{\oursRTEone}{0.9}   
\newcommand{\emdbVsBestWA}{-3.8\%}   
\newcommand{\emdbVsBestW}{+10.8\%}   
\newcommand{\emdbVsBestRTE}{-30.8\%} 
\newcommand{\richVsBestWA}{-17.8\%}   
\newcommand{\richVsBestW}{+23.1\%}    
\newcommand{\richVsBestRTE}{+22.7\%}  
\newcommand{\sceneVsBestAcc}{-5.1\%}  
\newcommand{\hthreeWA}{112.2}
\newcommand{\hthreeW}{267.9}
\newcommand{\hthreeRTE}{2.2}
\newcommand{\showWA}{109.1}\newcommand{\showW}{262.3}\newcommand{\showRTE}{2.1}
\newcommand{\showRichWA}{107.3}\newcommand{\showRichW}{172.7}\newcommand{\showRichRTE}{2.2}
\newcommand{\unishWA}{118.5}\newcommand{\unishW}{270.1}\newcommand{\unishRTE}{5.8}  

\newcommand{\slahmrWA}{326.9}\newcommand{\slahmrW}{776.1}\newcommand{\slahmrRTE}{10.2}
\newcommand{\whamWA}{135.6}\newcommand{\whamW}{354.8}\newcommand{\whamRTE}{6.0}
\newcommand{\gvhmrWA}{111.0}\newcommand{\gvhmrW}{276.5}\newcommand{\gvhmrRTE}{2.0}
\newcommand{\watchWA}{106.4}\newcommand{\watchW}{269.3}\newcommand{\watchRTE}{1.7}
\newcommand{\tramWA}{76.4}\newcommand{\tramW}{222.4}\newcommand{\tramRTE}{1.4}
\newcommand{\phmrWA}{71.0}\newcommand{\phmrW}{216.5}\newcommand{\phmrRTE}{1.3}
\newcommand{\joshWA}{68.9}\newcommand{\joshW}{174.7}\newcommand{\joshRTE}{1.3}      
\newcommand{\joshrWA}{220.0}\newcommand{\joshrW}{661.7}\newcommand{\joshrRTE}{13.1}  
\newcommand{\rteBestPrior}{1.3}      
\newcommand{\rteGain}{31\%}          

\newcommand{\ourFPSDeploy}{10.1}    
\newcommand{\benchGPU}{RTX PRO 6000 Blackwell}

\newcommand{\ablNoContactWA}{72.2}\newcommand{\ablNoContactW}{203.4}\newcommand{\ablNoContactRTE}{0.91}   
\newcommand{\noRulerWARatio}{2.7}      

\newcommand{\loraParams}{0.75M}
\newcommand{\floatBefore}{9.8\,cm}\newcommand{\floatAfter}{1.1\,cm}\newcommand{\penAfter}{1.2\%}
\newcommand{\skateVal}{1.45}\newcommand{\jitterVal}{3.9}
\newcommand{\rangeErrStart}{14.1\,cm}\newcommand{\rangeZoneStart}{100\%}
\newcommand{\ablNoRulerWA}{182.2}\newcommand{\ablNoRulerW}{1009.2}\newcommand{\ablNoRulerRTE}{11.3}  

\newcommand{\richSegs}{46}             
\newcommand{\richWA}{73.2}\newcommand{\richW}{163.1}\newcommand{\richRTEone}{2.7}

\newcommand{\richSigma}{7.2\%}         
\newcommand{\richCrowdSegs}{18}        
\newcommand{\richMaxPeople}{7}

\newcommand{\hthreeRichWA}{110.0}\newcommand{\hthreeRichW}{184.9}\newcommand{\hthreeRichRTE}{3.3}
\newcommand{\unishRichWA}{118.1}\newcommand{\unishRichW}{183.2}\newcommand{\unishRichRTE}{4.8}

\newcommand{\vggtLoraParams}{2.10M}      

\newcommand{\coinWA}{152.8}\newcommand{\coinW}{407.3}\newcommand{\coinRTE}{3.5}          
\newcommand{\tramRichWA}{127.8}\newcommand{\tramRichW}{238.0}\newcommand{\tramRichRTE}{6.0}   
\newcommand{\joshRichWA}{89.0}\newcommand{\joshRichW}{132.5}\newcommand{\joshRichRTE}{3.0}    

\makeatletter\newcommand{\captionof}[1]{\def\@captype{#1}\caption}\makeatother

\title{\fontsize{16.0pt}{18.9pt}\selectfont WildHSR: Metric Feed-Forward 4D People-Scene\\[2pt]Reconstruction from a 3D Foundation Model}

\author{Jerrin Bright, John Zelek \\
Vision and Image Processing Lab, University of Waterloo, Canada \\
{\tt\small \{j3bright, jzelek\}@uwaterloo.ca}}

\iclrfinalcopy
\begin{document}

\maketitle
\pagestyle{plain}
\thispagestyle{plain}
\begin{center}
\vspace{-1.2em}
\includegraphics[width=\textwidth]{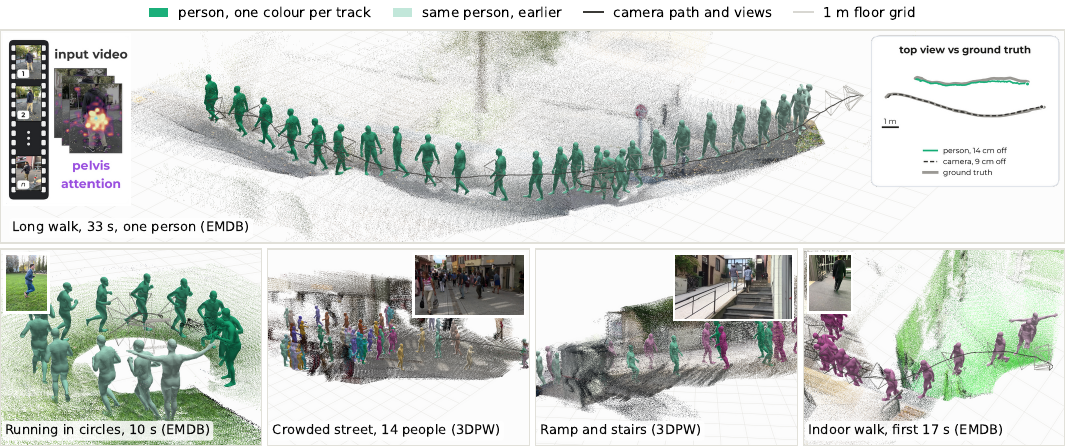}\vspace{6pt}
\captionof{figure}{\textbf{Metric people, scene and camera from one moving camera.} Each panel overlays persistent
people, the person-free scene and the camera path on a 1\,m grid. The large EMDB example spans 33\,s; its insets show
pelvis-token attention and a top view against ground truth after rigid rotation-and-translation alignment, preserving
the predicted scale. The remaining clips
cover cyclic motion, a 14-person street, stairs and an indoor walk. Inference receives a single monocular video.}
\label{fig:teaser}
\end{center}

\begin{abstract}
3D foundation models recover video cameras and geometry in one forward pass, but some of the strongest are up to
scale. Joint people-scene reconstruction then requires two missing outputs: \emph{metric scale} and persistent
\emph{person identity}. We ask whether one up-to-scale foundation representation can support both through
lightweight adaptation. Exact metric labels are scarce, but unlabeled in-the-wild video is abundant. We use people
in curated web video to initialise the solution: a posed metric body and 2D keypoints give an approximate,
closed-form scale pseudo-label.
These pseudo-labels pretrain a Scale Readout, which is then fine-tuned together with a lightweight adapter using
exact metric supervision from standard real-video training splits. At inference the head predicts metric scale
from foundation-model tokens, without the ruler or its teachers.
For \emph{person identity}, we probe the pretrained foundation model alone and find evidence that its intermediate
query-key features encode person correspondence across frames. \revision{In most evaluated moving-person clips,} a mid-layer token prefers that
person over the vacated location and other people. A tiny projection reads this correspondence; together with
metric pelvis motion and proposal confidence, it drives dustbin-aware Sinkhorn association of per-frame bodies.
\name{} combines both readouts to reconstruct metric cameras, scene and people from monocular video. Each window is
predicted feed-forward; analytic association and Sim(3) composition connect windows.
On EMDB-2, \name{} is the first feed-forward method in the published comparison to beat the best
optimization-based \revision{WA-MPJPE and RTE} while leading feed-forward methods on all three
world-frame metrics. On RICH, it leads feed-forward people-and-scene methods on WA-MPJPE and W-MPJPE.
\revision{The complete pipeline runs at} \ourFPSDeploy{} fps \revision{on one GPU.}
\end{abstract}

\section{Introduction}
\label{sec:intro}

Reconstructing people together with their surroundings from monocular video requires bodies, cameras and scene
geometry in one metric world. 3D foundation models now recover strong relative geometry and camera motion in a
single forward pass~\citep{dust3r,cut3r,vggt,vggtomega}, including video whose content moves~\citep{monst3r}.
Some learn metric output from metric supervision~\citep{cut3r,mapanything,metricanything}; some of the strongest
reported camera estimates instead come from models whose output is up to scale~\citep{vggt,vggtomega}, and it is
such a representation we study. Joint people-scene reconstruction then lacks two quantities. First, monocular
geometry is ambiguous up to a global \emph{metric scale}~\citep{eigen}, and an up-to-scale model normalises each
window to an arbitrary unit. Second, its geometric output has no persistent \emph{person identity}: it does not say
which body in one frame is the same person later. Over time, independently scaled windows distort trajectories, while identity switches splice different people into one path.

The prevailing answer is to obtain both outside the up-to-scale representation: metric scale from a metric-native backbone~\citep{human3r,cut3r}, from metric depth priors~\citep{tram}, or from synthetic metric
supervision and a depth teacher~\citep{unish,show}, and person identity from external detectors and
trackers~\citep{slahmr,wham,gvhmr}. These choices introduce separate models, supervision sources or sequence-processing stages. We ask whether one up-to-scale foundation representation can support both
capabilities without an external scale network or separately trained tracker. Scale requires supervision. Person
correspondence presents a different opportunity: 3D foundation models learn matching and geometric consistency
across views, so their intermediate representations may retain correspondence even on moving people. Large vision
models often encode structure beyond their explicit outputs: object segmentation emerges in the attention of
self-supervised transformers~\citep{dino}, 3D structure can be probed out of image foundation
models~\citep{probe3d}, and the attention of a pairwise geometry model separates moving objects from the static
scene with no training at all~\citep{easi3r}. We therefore transfer automatic
human-derived scale supervision into its scene tokens and test whether person correspondence can be read from its
pretrained representation~\citep{linearprobes}.

\begin{wrapfigure}{r}{0.49\textwidth}
\vspace{-\intextsep}
\centering
\includegraphics[width=\linewidth]{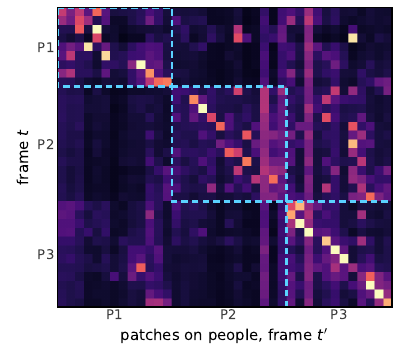}
\par\vspace{2pt}
\setlength{\abovecaptionskip}{0pt}
\footnotesize
\caption{\textbf{Person-patch matching at layer 13.} Rows are patches on three people at $t$;
columns are patches on the same people at $t'$. \currentrevision{Color shows pre-rotation query-key probe scores, normalized only over these candidates per head and then averaged, not scene-wide attention.} Dashed blocks mark same-person pairs; $82\%$ of top matches have the right identity here.}
\label{fig:intro_matching}
\end{wrapfigure}

\textbf{Reading person correspondence from the representation.} Reconstructing a body in each frame does not
establish which bodies belong to the same person over time. Rather than importing a separate visual tracker, we ask
whether the 3D backbone itself carries that correspondence. It learns geometric matching across views, yet moving
people are excluded from its matching supervision~\citep{vggtomega}. We probe the unmodified backbone before
training an identity readout and find \revision{a correspondence signal on people} in its intermediate query-key features.
Figure~\ref{fig:intro_matching} illustrates the finding: patches on each person in
one frame preferentially match patches on that person later, forming same-person blocks.
\revision{A retrieval probe validates the layer choice: intermediate query-key features retrieve
the person even after motion, while early and late features are nearer chance. This identifies a useful layer
for reading identity.} A small projection
extracts person features from the intermediate representation; metric pelvis motion and
proposal confidence then help resolve ambiguous links through dustbin-aware analytic assignment. Thus the same 3D
representation that supports people and scene reconstruction also supplies the correspondence cue for persistent
tracks, without a separately trained visual tracker.

\begin{figure}[t]
  \centering
  \includegraphics[width=\textwidth]{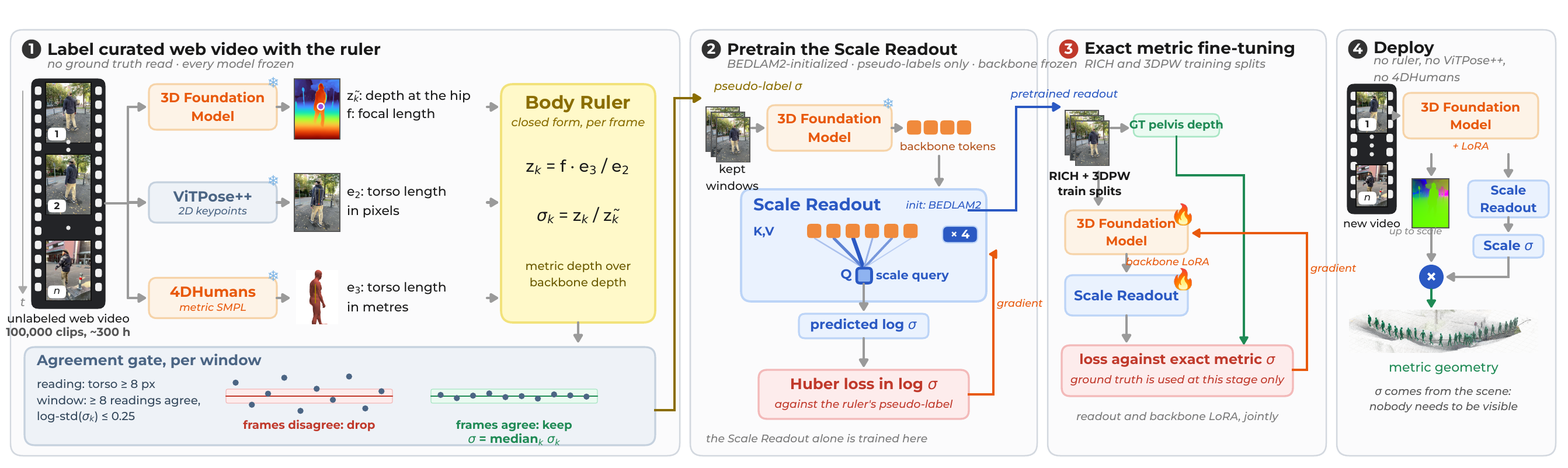}\vspace{6pt}
  \caption{\textbf{Training and deploying the Scale Readout.} (1)~A body ruler combines observed 2D torso extent
  $e_2$, the posed body's metric in-plane extent $e_3$, backbone depth $\tilde z_k$ and focal length $f$ to estimate
  $\sigma_k=(f e_3/e_2)/\tilde z_k$; an agreement gate yields window pseudo-labels.
  (2)~These labels pretrain the readout on web clips. (3)~Exact RICH and 3DPW targets fine-tune
  the readout and backbone LoRA. (4)~Deployment predicts $\sigma$ from backbone tokens alone,
  without the ruler or teachers.}
  \label{fig:ruler}
\end{figure}

\newpage
\textbf{Teaching metric scale to scene tokens.} Exact metric labels for ordinary real video are scarce. We instead
use people as offline rulers: a posed body supplies a metric torso extent in the image plane~\citep{smpl,smplx},
which we compare with 2D keypoints to obtain an approximate, closed-form scale pseudo-label (Fig.~\ref{fig:ruler}). Repeated readings expose
inconsistent labels, enabling pretraining on curated unlabeled web video before exact metric fine-tuning on standard
real-video training splits. The resulting Scale Readout predicts from scene-level foundation-model tokens rather
than repeating the body measurement at inference. This transfers human-derived supervision into a scale estimate
that is independent of a visible person at deployment.

\name{} is the system these two mechanisms make possible: metric cameras, scene and people from monocular video,
with no external scale network or separately trained external tracker (Fig.~\ref{fig:pipeline}). Each window is
reconstructed by a feed-forward network. Fixed analytic association and Sim(3) composition connect the window
predictions.
Our contributions are:
\begin{itemize}
  \item \textbf{\name{}}, \currentrevision{a system that jointly reconstructs people, cameras and scene geometry
  in one metric world from monocular video while maintaining person tracks.}
  \item \textbf{Emergent person correspondence in a 3D foundation model}: a controlled probe locates
  \revision{motion-robust identity cues in intermediate query-key features; a lightweight projection reads them
  for association without an external tracker} (\S\ref{sec:assoc}).
  \item \textbf{Human-derived scale supervision for scene tokens}: closed-form pseudo-labels from people in
  unlabeled video initialise a Scale Readout before exact metric adaptation (\S\ref{sec:exp_selfsup}).
  \item \textbf{\revision{State-of-the-art global motion}}: \revision{on EMDB-2, the first feed-forward method
  in our comparison to beat the best optimization method on WA-MPJPE and RTE; on RICH, the best feed-forward
  people-and-scene method on WA-MPJPE and W-MPJPE.}
\end{itemize}

\section{Related Work}
\label{sec:related}

\textbf{3D foundation models, and what they encode.} Feed-forward geometry models recover cameras and dense
structure from unposed images in one pass~\citep{dust3r,vggt}, online with a persistent state~\citep{cut3r} and in
the presence of motion~\citep{monst3r}. Those that output metres learn them from metric
supervision~\citep{cut3r,mapanything,metricanything}; some of the strongest reported camera estimates come from a model that is up to
scale and re-normalised per inference~\citep{vggt,vggtomega}. We build on that representation and learn its metric conversion. What such models encode beyond their outputs is less studied. Image foundation models carry
3D structure that probes can read~\citep{probe3d}, segmentation emerges in self-supervised attention~\citep{dino},
DUSt3R's attention separates moving from static content without training~\citep{easi3r}, and the authors of
VGGT-$\Omega$ report that clustering its intermediate tokens isolates a moving dancer, and that auxiliary
quantities, metric scale among them, can be decoded from its register tokens in preliminary
experiments~\citep{vggtomega}. We test two readouts: \emph{metric scale} from in-the-wild pseudo-labels followed by
exact adaptation, and \emph{person identity} from cross-frame correspondence in intermediate features.

\textbf{Supervision for metric scale.} Systems that reconstruct people and scenes in metres obtain metric scale from a metric-native backbone~\citep{human3r}, from metric depth priors~\citep{tram}, from synthetic metric supervision
with an expert depth teacher~\citep{unish}, or by training the body's scale prior into point-map
prediction~\citep{show}. Anthropometric size is an older cue. People as Scene Probes~\citep{sceneprobes} reads
depth, occlusion and lighting from passing pedestrians, SLAHMR and PACE use body priors inside their
objectives~\citep{slahmr,pace}, HAMSt3R~\citep{hamst3r} distils a mesh recovery encoder into a stereo network,
HSfM~\citep{hsfm} recovers approximate metric scale by per-scene optimization and, closest in mechanism,
HAC~\citep{hac} calibrates a SLAM reconstruction against the metric depth of contact joints from a mesh recovery
model. In these uses, the body measurement remains part of scale recovery at inference. We instead use the reading
offline to pretrain on curated unlabeled video, in the tradition of self-training~\citep{pseudolabel,noisystudent}
and consistency supervision~\citep{monodepth2}, then refine the predictor with exact metric labels from real-video
training splits. UniSH also trains on unlabeled in-the-wild video, but uses an external metric teacher; our
pseudo-label is produced by the geometric body ruler.

\textbf{Humans and scenes from video.} Global human mesh recovery places SMPL~\citep{smpl} bodies in world
coordinates, by optimization over SLAM and motion priors~\citep{slahmr,pace} or by regressing world
trajectories~\citep{glamr,wham,tram,gvhmr,prompthmr}. Closest to us, Human3R~\citep{human3r}, UniSH~\citep{unish},
SHOW~\citep{show} and GUSH3R~\citep{gush3r} attach human decoders to geometry foundation models and reconstruct
people and scene in one feed-forward pass, the last as Gaussians; MetricHMSR~\citep{metrichmsr} does so metrically
from one image, and JOSH3R~\citep{josh} is trained from the pseudo-labels of a per-sequence optimization. These
systems differ in where the human branch reads the backbone. Human3R decodes SMPL-X parameters at a detected head
cell, which becomes ambiguous when two people share that token. \revision{We read each person at
the pelvis, a geometric anchor for body placement} (\S\ref{sec:assoc}).

\section{Method}
\label{sec:method}

\subsection{Problem formulation and system overview}
\label{sec:problem}
\begin{figure}[t]
  \centering
  \includegraphics[width=\textwidth]{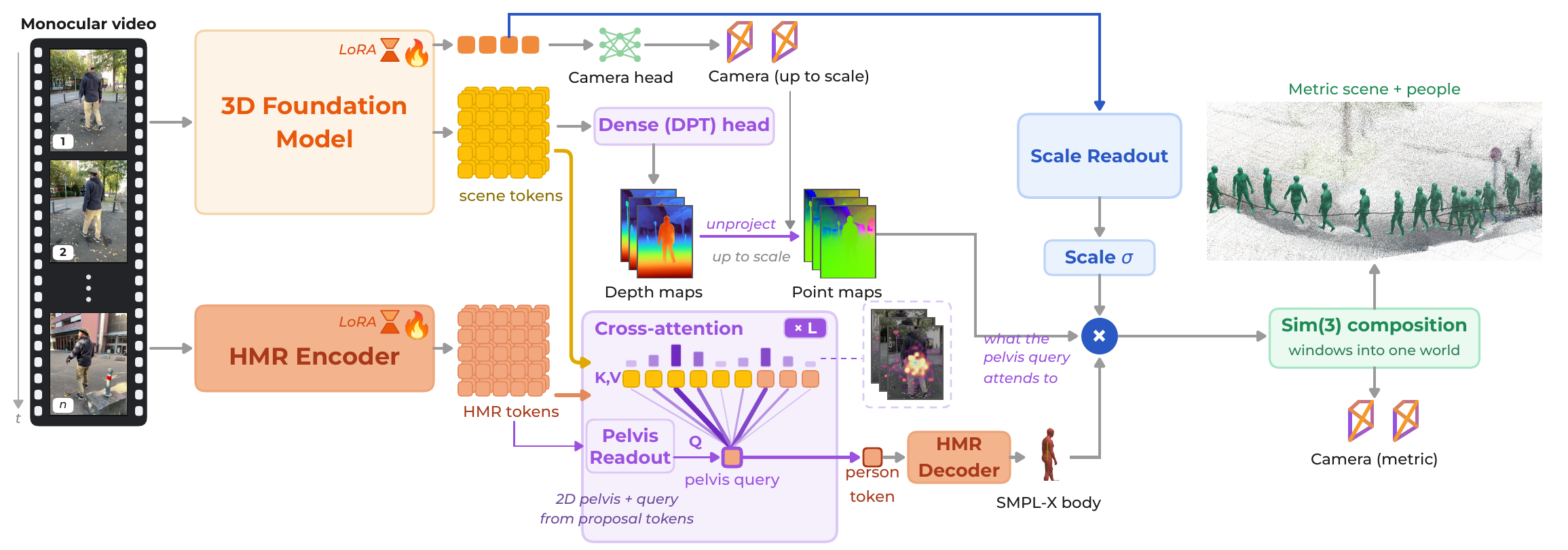}\vspace{6pt}
  \caption{Overview of \textbf{\name{}.} A 3D foundation model produces up-to-scale cameras, geometry and scene tokens.
  A mesh branch proposes people; \revision{a Pelvis Readout forms a body-localized query}, and cross-attention
  decodes a metric SMPL-X body.
  The Scale Readout supplies $\sigma$; analytic association and Sim(3) composition link
  people and windows in one metric world.}
  \label{fig:pipeline}
\end{figure}

A window of video passes once through a 3D foundation model (VGGT-$\Omega$~\citep{vggtomega}, the successor of
VGGT~\citep{vggt}), adapted with a small LoRA for metric transfer, which
predicts per-frame cameras with focal length $f$ and depth $\tilde z$, from which point maps
are obtained by unprojection, together with the tokens used to decode these outputs
(Fig.~\ref{fig:pipeline}, top). All of it is correct up to one unknown
scale. We write $\sigma = z_{\text{metric}}/z_{\text{backbone}}$ for the metres-per-unit
conversion of that inference. The backbone re-normalises every pass, so $\sigma$ must be predicted per window rather
than pooled over a sequence. Cameras and scene are in backbone units, whereas SMPL-X dimensions are metric;
$\sigma$ \revision{puts both in one world.}

\revision{The body prior supplies shape, not placement; VGGT supplies cameras and depth, not identity.
Cross-attention grounds proposals; scale and association connect windows.}

Two pretrained networks supply scene and human tokens, while our Scale Readout, \revision{Pelvis Readout}, cross-attention
fusion and identity projection expose the quantities needed for joint reconstruction. Their released weights remain frozen; small LoRA
adapters support metric and body adaptation. Appendix~\ref{app:compute} gives the exact modules, parameter counts and
training configuration. We next describe how the system reads and associates \textbf{people}, learns \textbf{scale},
and composes one metric \textbf{world}.

\needspace{12\baselineskip}
\subsection{Person reconstruction and temporal association}
\label{sec:assoc}
\label{sec:crowd}
\textbf{Per-frame proposals and bodies.} Multi-HMR~\citep{multihmr}, adapted with a
\loraParams{} LoRA, scores person-centre patches and proposes initial SMPL-X bodies per frame; VGGT-$\Omega$
\revision{independently supplies scene tokens for the video window. For proposal $n$ at time $t$, a lightweight
Pelvis Readout takes its HMR tokens and produces a 2D pelvis location and person query. During training, the
location is supervised by the pelvis joint of the ground-truth SMPL-X body projected into the image; the target
does not come from the model's own final prediction. We then fuse the readout query with the two token streams:}
\begin{equation}
\begin{aligned}
 (u_{t,n},q^{\rm pel}_{t,n})&=R_{\rm pel}(H_{t,n}),\\
 h_{t,n}&=\operatorname{CrossAttn}\!\left(q^{\rm pel}_{t,n},[S_{1:T};H_t]\right),\quad
 B_{t,n}=D_{\rm HMR}(h_{t,n}).
\end{aligned}
\label{eq:person-fusion}
\end{equation}
Here $H_{t,n}$ denotes the HMR tokens of proposal $n$, $u_{t,n}$ is the readout's image-space pelvis location,
$q^{\rm pel}_{t,n}$ its query, $S_{1:T}$ the VGGT scene tokens, and $H_t$ the frame's HMR tokens;
\revision{the latter two supply keys and values. At inference, the Pelvis Readout needs no ground-truth body or
external keypoint model.}
\revision{No external detector runs in the deployed path.}

\textbf{The pelvis anchor.} \revision{Placement unprojects the Pelvis Readout's pixel} $u_{t,n}$ \revision{using the median backbone
depth} $\tilde z_{t,n}$ \revision{over the body's projected torso.} With backbone-unit camera centre $C_t$,
camera-to-window rotation $R_t$, intrinsics $K_t$ and $\bar u_{t,n}=[u_{t,n}^{\mathsf T},1]^{\mathsf T}$, \revision{the metric pelvis in window $w$ is}
\begin{equation}
 p^{(w)}_{t,n}=\sigma\bigl(C_t+R_t(\tilde z_{t,n}K_t^{-1}\bar u_{t,n})\bigr).
 \label{eq:pelvis-placement}
\end{equation}
\revision{The readout is supervised at SMPL's hip midpoint, so} $p^{(w)}_{t,n}$ \revision{places the body's root. The torso
median avoids a body--ground depth discontinuity at the anchor. Direct metric-translation regression performs
worse; a head anchor introduces an orientation-sensitive lever arm of roughly} $0.6$\,m.

\FloatBarrier
\begin{figure}[t]
\centering
\includegraphics[width=\textwidth]{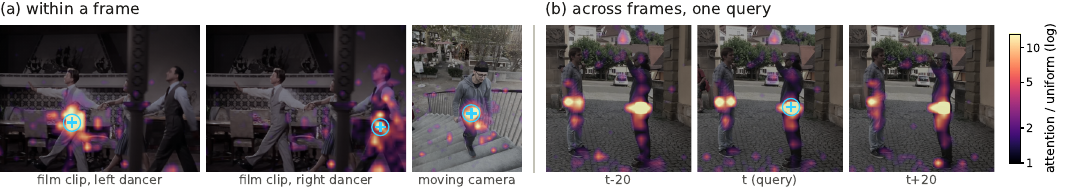}\vspace{8pt}
\caption{\textbf{Pretrained person correspondence.} At layer 13, a VGGT-$\Omega$ pelvis token attends to its own
person within a frame ($3.8\times$ uniform over four clips) and across the window ($5.8\times$). \revision{When the
person moves,} attention follows the person ($5.0\times$), not the vacated location ($1.2\times$). This probe runs
the backbone alone; Appendix~\ref{app:identity} gives controls and the complete sample accounting.}
\label{fig:attn}
\end{figure}

\textbf{Identity readout and association.} The probe of Fig.~\ref{fig:attn} peaks in the layer-13 query-key space.
We pool those vectors around each pelvis and map them to a 128-dimensional unit descriptor with a two-layer
projection trained by supervised contrastive loss. For track $i$ and proposal $j$, let $m_i,e_j$ be unit
descriptors, $\hat p_i,p_j$ their predicted and observed metric pelvis positions, $c_j$ proposal confidence and
$g_i$ a time-dependent motion gate. \revision{Their assignment cost is}
\begin{equation}
 C_{ij}=\tfrac12(1-\langle m_i,e_j\rangle)
 +\min\!\left(\frac{\lVert p_j-\hat p_i\rVert_2}{g_i},1\right)
 +0.25(1-c_j).
 \label{eq:person-association}
\end{equation}
\revision{Pairs beyond} $g_i$ \revision{are invalid;} Appendix~\ref{app:identity} \revision{specifies the gate and track state.} A dustbin and Sinkhorn
optimal transport~\citep{sinkhorn} give a soft assignment, which is hardened one-to-one with Hungarian matching.
The projection also receives a ground-truth assignment loss through the soft Sinkhorn matrix; Hungarian selection
and confidence-gated memory updates remain outside backpropagation. Appendix~\ref{app:identity} gives the full
specification. Learned inference is feed-forward within each window; \revision{fixed-step association and memory updates
require no test-time gradient-based fitting.}

\subsection{Scale Readout and metric adaptation}
\label{sec:selfsup}
\label{sec:ruler}

\textbf{\revision{Ruler pseudo-labels.}} \revision{For each visible person,} ViTPose++-H~\citep{vitposepp} at
$256\times192$ gives the observed image-plane extent $e_2$ from shoulder midpoint to hip midpoint.
4DHumans~\citep{hmr2} supplies the posed metric torso; $e_3$ is the in-plane extent between its
corresponding midpoints. With the backbone's focal length $f$ and estimated depth
$\tilde z_k$, a local weak-perspective approximation gives the ruler reading
(Fig.~\ref{fig:ruler}, stage 1)
\begin{equation}
  z_k = f\,\frac{e_3}{e_2}, \qquad \sigma_k = \frac{z_k}{\tilde z_k},
  \label{eq:ruler}
\end{equation}
without a metric label or sensor for the web clip. The estimate can be biased when the torso endpoints
have different depths. We retain confident detections with $e_2\ge8$\,px; a window needs at least
eight readings and $\mathrm{std}(\log\sigma_k)\le0.25$. Its pseudo-label is the median across eligible
frames and people; agreement tests consistency, not absolute correctness.

\textbf{\revision{Pretraining and adaptation.}} \revision{These labels pretrain the Scale Readout on 100,000 curated
web-video clips (about 300 hours). The clips include publicly accessible video such as YouTube; YOLO person counts
guide sampling across crowd sizes. A learned query reads the backbone's camera and register tokens to regress}
$\log\sigma$ \revision{with a log-space Huber loss while the backbone remains fixed}
(architecture in Appendix~\ref{app:compute}).
\revision{After synthetic initialization and ruler pretraining, we fine-tune the readout and backbone LoRA with exact}
metric targets from standard RICH~\citep{rich} and 3DPW~\citep{3dpw} training splits~\citep{human3r}.
\revision{Camera calibration and body translation provide reference pelvis depth; no scene point cloud is needed.
At inference, the readout needs only backbone tokens, with no ruler, teacher or visible person.}

\subsection{Composing one metric world}
\label{sec:stitch}
Given $\sigma$, the backbone's camera centres, depth and point maps are multiplied by it, while the
metric body is not, and one forward pass per window yields the metric scene, the metric camera
path and the people placed in it (Fig.~\ref{fig:pipeline}). Long video needs one continuous world
frame, but every window is an independent inference with its own frame and unit. We use 100-frame windows with stride
50 and require at least eight shared frames. \needspace{6\baselineskip}For consecutive windows $a$ and $b$, the relative rotation $R_{ba}$ is
the chordal mean of the framewise rotations between their shared camera orientations. With $R_{ba}$ fixed, we solve
\begin{equation}
(s_{ba},t_{ba})=\arg\min_{s,t}\sum_{k\in a\cap b}\left\|C^a_k-(sR_{ba}C^b_k+t)\right\|_2^2,
\label{eq:window-compose}
\end{equation}
in closed form on their metric camera centres. If fewer than eight correspondences survive or the centred camera
trajectory is degenerate, we retain the previous cumulative similarity. All shared frames have equal weight and no
outlier trimming is applied. We compose valid relative similarities in temporal order. If $S_w$ is the
cumulative scale of window $w$, we divide every cumulative similarity by
$g=\exp\!\left(|\mathcal W|^{-1}\sum_w\log S_w\right)$, preserving the sequence's aggregate metric scale rather
than letting one window set it. The similarity acts fully on camera centres, point maps and pelvis translations;
its rotation also acts on camera and SMPL-X global orientations, while body dimensions remain unscaled and no
network or scene parameters are optimized at test time.

\begin{figure}[t]
\centering
\includegraphics[width=0.92\textwidth]{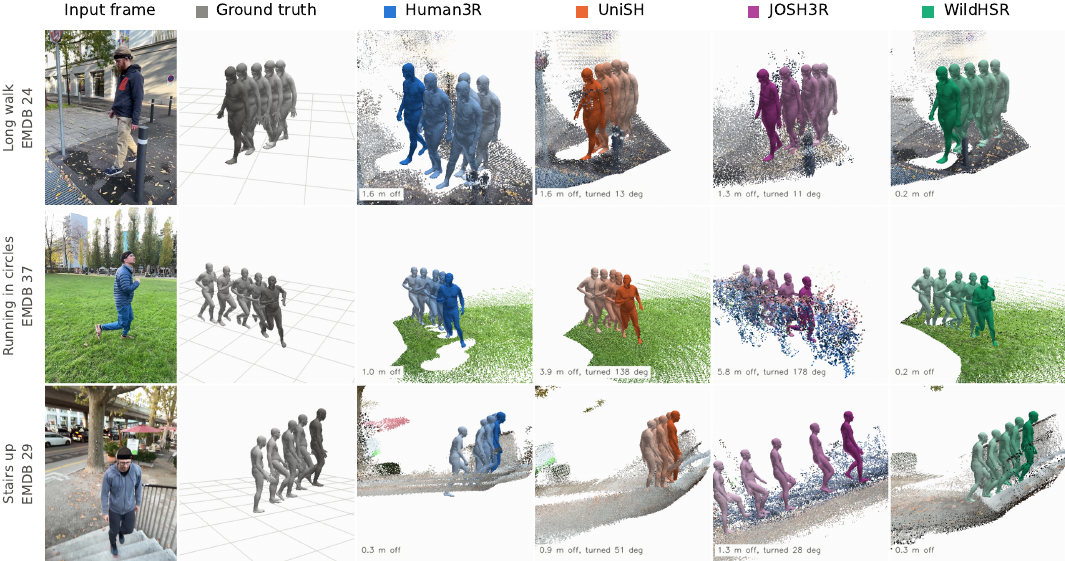}
\setlength{\abovecaptionskip}{3pt}
\caption{\textbf{Scene and people in one metric world} (three EMDB-2 clips), each method at its own metric scale; labels
give placement error and, for UniSH and JOSH3R, view turn. \name{} is $0.2$ to $0.3$\,m off, against $0.3$ to $1.6$\,m
(Human3R), $0.9$ to $3.9$\,m (UniSH) and $1.3$ to $5.8$\,m (JOSH3R).}
\label{fig:qual_main}
\end{figure}

\needspace{7\baselineskip}
\subsection{Training objectives}
\label{sec:training}
The scale pathway uses the log-space Huber objective above; \revision{the Pelvis Readout's location is supervised
by projected ground-truth SMPL-X pelvis joints, while its query and the fusion receive gradients through the decoded
body. No target is imposed on the cross-attention weights themselves.} Body losses cover SMPL-X parameters, mesh and
reprojection; identity losses use contrastive and soft assignment, while \currentrevision{scene contact regularizes placement.}

\textbf{Feet on the reconstructed ground.} \revision{A scene-contact consistency term encourages supporting feet
to agree with the local reconstructed surface during training. The body branch predicts its placement directly at
inference, without a manual vertical shift} (Appendix~\ref{app:ablations}).

\FloatBarrier

\section{Experiments}
\label{sec:experiments}

\textbf{Protocol.} We follow the published EMDB-2~\citep{emdb} 25-sequence and RICH~\citep{rich} protocols used by
the compared methods~\citep{human3r,unish,show,josh,watch,trace,coin}. WA-MPJPE uses similarity alignment,
W-MPJPE aligns the first two frames, and RTE aligns trajectories by rotation and translation only; RTE therefore
preserves scale error.

\begin{table}[t]
\centering
\small
\begin{minipage}[t]{0.55\textwidth}
\centering
\caption{\textbf{EMDB-2 global motion.} \currentrevision{WA/W: global joint error; RTE: trajectory error.} \ranklegendF{} \currentrevision{applies throughout.}}
\label{tab:main}
\setlength{\tabcolsep}{4pt}
\begin{tabular}{@{}clccc@{}}
\toprule
& method & WA $\downarrow$ & W $\downarrow$ & RTE $\downarrow$ \\
\midrule
\multirow{6}{*}{\rotatebox[origin=c]{90}{\scriptsize Optimization-based}}
& SLAHMR           & \slahmrWA{} & \slahmrW{} & \slahmrRTE{} \\
& COIN               & \coinWA{}   & \coinW{}   & \coinRTE{} \\
& TRAM               & \tramWA{} & \tramW{} & \thirdc{\tramRTE{}} \\
& PromptHMR-vid & \thirdc{\phmrWA{}} & \thirdc{\phmrW{}} & \secondc{\phmrRTE{}} \\
& JOSH               & \secondc{\joshWA{}} & \bestc{\joshW{}} & \secondc{\joshRTE{}} \\
\midrule
\multirow{8}{*}{\rotatebox[origin=c]{90}{\scriptsize Feed-forward}}
& WHAM               & \whamWA{}   & \whamW{}   & \whamRTE{} \\
& GVHMR             & \gvhmrWA{}  & \gvhmrW{}  & \gvhmrRTE{} \\
& WATCH             & \watchWA{}  & \watchW{}  & \watchRTE{} \\
& JOSH3R & \joshrWA{} & \joshrW{} & \joshrRTE{} \\
& Human3R         & \hthreeWA{} & \hthreeW{} & \hthreeRTE{} \\
& UniSH             & \unishWA{}  & \unishW{}  & \unishRTE{} \\
& SHOW               & \showWA{}   & \showW{}   & \showRTE{} \\
& \textbf{\name{}} & \bestc{\oursWA{}} & \secondc{\oursW{}} & \bestc{\oursRTEone{}} \\
\cmidrule(lr){2-5}
& \scriptsize $\Delta$ vs best & \scriptsize \emdbVsBestWA{} & \scriptsize \emdbVsBestW{} & \scriptsize \emdbVsBestRTE{} \\
\bottomrule
\end{tabular}
\end{minipage}\hfill
\begin{minipage}[t]{0.42\textwidth}
\centering
\caption{\textbf{RICH global motion.} \currentrevision{WA/W: global joint error; RTE: per-segment trajectory error.}}
\label{tab:rich}
\setlength{\tabcolsep}{4pt}
\begin{tabular}{lccc}
\toprule
method & WA $\downarrow$ & W $\downarrow$ & RTE $\downarrow$ \\
\midrule
\multicolumn{4}{l}{\emph{Optimization-based}} \\
TRAM     & \tramRichWA{} & \tramRichW{} & \tramRichRTE{} \\
JOSH     & \secondc{\joshRichWA{}} & \bestc{\joshRichW{}} & \thirdc{\joshRichRTE{}} \\
\midrule
\multicolumn{4}{l}{\emph{Feed-forward, people and scene}} \\
Human3R  & \hthreeRichWA{} & \hthreeRichW{} & \hthreeRichRTE{} \\
UniSH    & \unishRichWA{} & \unishRichW{} & \unishRichRTE{} \\
SHOW     & \thirdc{\showRichWA{}} & \thirdc{\showRichW{}} & \bestc{\showRichRTE{}} \\
\textbf{\name{}} & \bestc{\richWA{}} & \secondc{\richW{}} & \secondc{\richRTEone{}} \\
\cmidrule(lr){1-4}
\scriptsize $\Delta$ vs best & \scriptsize \richVsBestWA{} & \scriptsize \richVsBestW{} & \scriptsize \richVsBestRTE{} \\
\bottomrule
\end{tabular}
\end{minipage}

\vspace{2pt}
\vspace{2pt}
\caption{\textbf{Training-data ablation on EMDB-2.} \currentrevision{End-to-end joint-position (WA, W) and trajectory (RTE) errors for checkpoints trained with the indicated data sources.}}
\label{tab:scale_stages}
\label{tab:selfsup}
\setlength{\tabcolsep}{7pt}
\renewcommand{\arraystretch}{0.94}
\begin{tabular}{lccc@{\hspace{1.8em}}ccc}
\toprule
& \multicolumn{3}{c}{training data} & \multicolumn{3}{c}{EMDB-2} \\
\cmidrule(lr){2-4}\cmidrule(lr){5-7}
checkpoint & BEDLAM2 & curated ITW & RICH+3DPW & WA $\downarrow$ & W $\downarrow$ & RTE $\downarrow$ \\
\midrule
synthetic base & \cmark & & & \ablNoRulerWA{} & \ablNoRulerW{} & \ablNoRulerRTE{} \\
ITW pretrained & \cmark & \cmark & & \thirdc{136.7} & \thirdc{418.6} & \thirdc{4.63} \\
no ITW pretraining & \cmark & & \cmark & \secondc{116.3} & \secondc{255.9} & \secondc{3.7} \\
\textbf{metric fine-tuned} & \cmark & \cmark & \cmark & \bestc{\oursWA{}} & \bestc{\oursW{}} & \bestc{\oursRTE{}} \\
\bottomrule
\end{tabular}

\vspace{5pt}
\noindent\begin{minipage}[t]{0.48\textwidth}
\vspace{0pt}
\centering
\caption{\textbf{Held-out 3DPW body reconstruction.} 14-joint test errors in mm; published baselines from SHOW~\citep{show}.}
\label{tab:3dpw_local_published}
\scriptsize
\setlength{\tabcolsep}{2pt}
\begin{tabular}{lccc}
\toprule
method & PA-MPJPE$\downarrow$ & MPJPE$\downarrow$ & PVE$\downarrow$ \\
\midrule
Human3R & \thirdc{44.1} & \thirdc{71.2} & \thirdc{84.9} \\
UniSH & 48.8 & 75.6 & 88.8 \\
SHOW & \secondc{41.0} & \secondc{67.7} & \secondc{78.7} \\
\textbf{WildHSR} & \bestc{38.7} & \bestc{65.2} & \bestc{74.4} \\
\bottomrule
\end{tabular}
\end{minipage}\hfill
\begin{minipage}[t]{0.48\textwidth}
\vspace{0pt}
\small
\textbf{Local body accuracy.} Table~\ref{tab:3dpw_local_published} evaluates held-out 3DPW. PA-MPJPE removes per-pose similarity; MPJPE and PVE measure camera-frame joints and mesh. \name{} improves on SHOW, the strongest baseline, by $2.3$, $2.5$ and $4.3$\,mm, respectively. These metrics do not assess world placement. Together with Tables~\ref{tab:main} and~\ref{tab:rich}, they show that \name{} gains global accuracy without sacrificing local body fidelity.
\end{minipage}
\end{table}

\subsection{Metric cameras, scene and people}
\label{sec:exp_main}
\label{sec:bodyscale}
\textbf{EMDB-2.} Table~\ref{tab:main} \currentrevision{compares} \name{} \currentrevision{with previously reported EMDB-2 results.} \emph{Trajectory error is the best published}:
RTE \oursRTEone{} improves on the
prior best (\rteBestPrior{}) by \rteGain{}. On the joint metrics only JOSH~\citep{josh}, which optimizes each sequence jointly over scene and body, is ahead on
W-MPJPE (\joshW{} against \oursW{}). \name{} has the lowest WA-MPJPE in the full table (\oursWA{}).
\emph{Among feed-forward methods \name{} leads all three metrics}, by $38\%$, $26\%$ and $47\%$ over the best
of the others on each. What differs is what each system spends to become
metric: metric depth priors, large-scale supervised pretraining or a metric-native
backbone, where \name{} learns scale from the people.
Fig.~\ref{fig:qual_main} shows the reconstructions behind these numbers, and Appendix~\ref{app:qual} the
paths themselves (Fig.~\ref{fig:traj}) and six more clips. The visual comparison is restricted to systems whose
neural reconstruction networks jointly predict people and scene:
Human3R, UniSH and JOSH3R~\citep{josh}; SHOW~\citep{show} has not released weights and appears in the tables only
(alignment and re-run details in Appendix~\ref{app:qual}).
\currentrevision{Because RTE preserves scale error, this comparison tests the shared metric scene and pelvis-based
body placement, not only local pose quality.}

\textbf{RICH.} \name{} has the best WA-MPJPE of Table~\ref{tab:rich} and leads the feed-forward people-and-scene methods on
W-MPJPE, where only JOSH, an optimization, is ahead. Its median multiplicative scale error
is \richSigma{}. Its per-segment RTE is second among feed-forward people-and-scene systems
(\richRTEone{}), behind SHOW at \showRichRTE{}; subjects are near-stationary and RTE divides by a
median displacement of $0.52$\,m. \revision{Our model is fine-tuned with exact labels from RICH-train and 3DPW-train;
the RICH test partition is held out.} Table~\ref{tab:rich} uses the same test split and evaluation protocol as the
baselines, \revision{not necessarily the same training data.} \currentrevision{With camera motion largely removed,
the joint results are consistent with the Scale Readout and pelvis-to-scene fusion placing bodies in a shared metric frame.}

\textbf{Pretrained person correspondence.}\label{sec:exp_identity} Figures~\ref{fig:intro_matching}
and~\ref{fig:attn} \revision{show intermediate features matching a moving person across frames. Tested on 1,500 clips,}
same-person retrieval reaches $80$ to $82\%$ in layers 11 to 15 against $46\%$ chance. \revision{This retrieval accuracy is
distinct from the conditional match weights in} Fig.~\ref{fig:intro_matching} \revision{and the attention lift in}
Fig.~\ref{fig:attn}. \currentrevision{The identity projection reads these mid-depth features and combines them with
metric pelvis motion for cross-frame assignment} (Appendix~\ref{app:identity}).

\textbf{Long video and runtime.} Over 1000 frames (33\,s), \name{}'s camera-to-person distance error is $5.4$ to $6.0\%$
($9.4$ to $12.9\%$ for Human3R); without alignment, its median range error is \rangeErrStart{} and its proxemic
zone is correct on \rangeZoneStart{} of frames. The complete pipeline runs at \ourFPSDeploy{} fps on one
\benchGPU{}. \currentrevision{Fixed association and Sim(3) composition link feed-forward windows without per-video
fitting}; Appendix~\ref{app:compute} gives the matched runtime breakdown.

\begin{table}[H]
\centering
\small
\caption{\textbf{Association on 13 two-person 3DPW test sequences.}
WildHSR variants share fixed Multi-HMR outputs; Human3R uses its native pipeline.
Assignment thresholds were tuned on four validation sequences.}
\label{tab:identity_assoc}
\scriptsize
\setlength{\tabcolsep}{4.2pt}
\begin{tabular}{lccccc}
\toprule
method & IDF1$\uparrow$ & HOTA$\uparrow$ & ID switches$\downarrow$ & fragments$\downarrow$ & MOTA$\uparrow$ \\
\midrule
Human3R (native) & \bestc{87.4} & \secondc{72.3} & 1682 & \thirdc{87} & \thirdc{87.1} \\
\midrule
motion + confidence DP & \thirdc{66.9} & \thirdc{56.9} & \secondc{97} & \secondc{77} & \secondc{87.5} \\
raw VGGT query/key & 51.8 & 53.4 & 8065 & 861 & 51.7 \\
projected identity & 60.2 & 54.5 & \thirdc{664} & 284 & 73.7 \\
\textbf{projection + motion + confidence} & \secondc{82.1} & \bestc{74.5} & \bestc{39} & \bestc{73} & \bestc{96.5} \\
\bottomrule
\end{tabular}
\end{table}

\textbf{Temporal association.} Table~\ref{tab:identity_assoc} compares association cues on 13 two-person 3DPW
test sequences with fixed Multi-HMR proposals and bodies. The combined cost reaches $74.5$ HOTA and $39$ ID
switches, versus $72.3$ and $1682$ for native Human3R. Human3R leads IDF1 ($87.4$ versus $82.1$) and nine
sequences; one crowded clip dominates its switch count. The weaker single-cue variants support combining
projected identity with motion, but do not isolate confidence. This controlled component test does not measure the final end-to-end pipeline (Appendix~\ref{app:association}).

\subsection{Effect of the scale-training stages}
\label{sec:exp_selfsup}
Table~\ref{tab:scale_stages} \revision{tests web-video pseudo-labeling within the final recipe. Removing it while
keeping BEDLAM2 initialization and exact RICH/3DPW fine-tuning worsens WA/W/RTE from}
\oursWA{}/\oursW{}/\oursRTE{} \revision{to 116.3/255.9/3.7. The pseudo-label stage contributes beyond exact labels.}

\section{Conclusion}
\label{sec:conclusion}

\name{} transfers human-derived scale pseudo-labels into an up-to-scale 3D foundation model and reads person
correspondence from its pretrained intermediate query-key features. The Scale Readout is initialized on real-video
pseudo-labels before exact metric adaptation; a small identity projection combines the correspondence signal with
metric motion for association. Together they enable joint metric reconstruction of people, cameras and scene in
feed-forward windows, linked by analytic association and Sim(3) composition \revision{without test-time optimization}.

\textbf{Limitations.} \revision{Identity weakens under interaction and occlusion, and its specificity to people is untested.
Sequence composition is offline; RICH training mixtures may differ across methods.}

\section*{Reproducibility statement}
VGGT-$\Omega$ and Multi-HMR retain their released base weights; small LoRAs adapt them~\citep{lora}. The Scale
Readout starts on BEDLAM2, learns from offline ruler pseudo-labels, then receives exact metric supervision from
RICH- and 3DPW-train; EMDB-2 is held out. The ruler, agreement gate, ViTPose++-H and 4DHumans are absent at
inference. The appendix details objectives, splits, settings and evaluation; code, models and outputs will be released.

\section*{Ethics and web-data statement}
Public web video supplies 100,000 clips for non-identifying scale pseudo-labels; we do not redistribute it.
Persistent reconstruction may enable surveillance. Track IDs do not identify people, and the system should not
be used for biometric or high-stakes decisions. Web video and parametric body priors may introduce demographic,
body-shape, clothing, mobility and visibility biases.

\section*{AI use statement}
Large language models assisted manuscript drafting, restructuring and editing. The authors directed and revised
the generated text and take full responsibility for this paper.

\clearpage
\bibliography{refs}

@article{human3r,
  title   = {Human3R: Everyone Everywhere All at Once},
  author  = {Chen, Yue and Chen, Xingyu and Xue, Yuxuan and Chen, Anpei and Xiu, Yuliang and Pons-Moll, Gerard},
  journal = {arXiv preprint arXiv:2510.06219},
  year    = {2025},
}

@inproceedings{cut3r,
  title     = {Continuous 3{D} Perception Model with Persistent State},
  author    = {Wang, Qianqian and Zhang, Yifei and Holynski, Aleksander and Efros, Alexei A.
               and Kanazawa, Angjoo},
  booktitle = {IEEE/CVF Conference on Computer Vision and Pattern Recognition (CVPR)},
  year      = {2025},
}

@inproceedings{vggt,
  title     = {VGGT: Visual Geometry Grounded Transformer},
  author    = {Wang, Jianyuan and Chen, Minghao and Karaev, Nikita and Rupprecht, Christian and Novotny, David},
  booktitle = {CVPR},
  year      = {2025}
}

@inproceedings{multihmr,
  title     = {Multi-HMR: Multi-Person Whole-Body Human Mesh Recovery in a Single Shot},
  author    = {Baradel, Fabien and Armando, Matthieu and Galaaoui, Salma and Br{\'e}gier, Romain
               and Weinzaepfel, Philippe and Rogez, Gr{\'e}gory and Lucas, Thomas},
  booktitle = {ECCV},
  year      = {2024}
}

@inproceedings{sinkhorn,
  title     = {Sinkhorn Distances: Lightspeed Computation of Optimal Transport},
  author    = {Cuturi, Marco},
  booktitle = {Advances in Neural Information Processing Systems},
  year      = {2013}
}

@inproceedings{smpl,
  title     = {{SMPL}: A Skinned Multi-Person Linear Model},
  author    = {Loper, Matthew and Mahmood, Naureen and Romero, Javier and Pons-Moll, Gerard and Black, Michael J.},
  booktitle = {ACM TOG (SIGGRAPH Asia)},
  year      = {2015}
}

@inproceedings{emdb,
  title     = {{EMDB}: The Electromagnetic Database of Global 3D Human Pose and Shape in the Wild},
  author    = {Kaufmann, Manuel and Song, Jie and Guo, Chen and Shen, Kaiyue and Jiang, Tianjian
               and Tang, Chengcheng and Zarate, Juan and Hilliges, Otmar},
  booktitle = {ICCV},
  year      = {2023}
}

@inproceedings{slahmr,
  title     = {Decoupling Human and Camera Motion from Videos in the Wild},
  author    = {Ye, Vickie and Pavlakos, Georgios and Malik, Jitendra and Kanazawa, Angjoo},
  booktitle = {CVPR},
  year      = {2023}
}

@inproceedings{pace,
  title     = {{PACE}: Human and Camera Motion Estimation from in-the-wild Videos},
  author    = {Kocabas, Muhammed and Yuan, Ye and Molchanov, Pavlo and Guo, Yunrong and Black, Michael J.
               and Hilliges, Otmar and Kautz, Jan and Iqbal, Umar},
  booktitle = {3DV},
  year      = {2024}
}

@inproceedings{glamr,
  title     = {{GLAMR}: Global Occlusion-Aware Human Mesh Recovery with Dynamic Cameras},
  author    = {Yuan, Ye and Iqbal, Umar and Molchanov, Pavlo and Kitani, Kris and Kautz, Jan},
  booktitle = {CVPR},
  year      = {2022}
}

@inproceedings{wham,
  title     = {{WHAM}: Reconstructing World-grounded Humans with Accurate 3D Motion},
  author    = {Shin, Soyong and Kim, Juyong and Halilaj, Eni and Black, Michael J.},
  booktitle = {CVPR},
  year      = {2024}
}

@inproceedings{tram,
  title     = {{TRAM}: Global Trajectory and Motion of 3D Humans from in-the-wild Videos},
  author    = {Wang, Yufu and Wang, Ziyun and Liu, Zhicheng and Daniilidis, Kostas},
  booktitle = {ECCV},
  year      = {2024}
}

@misc{unish,
  title  = {{UniSH}: Unifying Scene and Human Reconstruction in a Feed-Forward Pass},
  author = {Li, Mengfei and Li, Peng and Zhang, Zheng and Lu, Jiahao and Zhao, Chengfeng
            and Xue, Wei and Liu, Qifeng and Peng, Sida and Zhang, Wenxiao and Luo, Wenhan
            and Liu, Yuan and Guo, Yike},
  year   = {2026},
  eprint = {2601.01222},
  archivePrefix = {arXiv}
}

@misc{show,
  title  = {Scene and Human in One World: Reconstruction in a Feedforward Pass},
  author = {Shi, Boao and Feng, Qiao and Huang, Yiming and Liu, Lingjie},
  year   = {2026},
  eprint = {2606.27720},
  archivePrefix = {arXiv},
}

@inproceedings{hamst3r,
  title     = {{HAMSt3R}: Human-Aware Multi-view Stereo 3{D} Reconstruction},
  author    = {Rojas, Sara and Armando, Matthieu and Ghanem, Bernard and Weinzaepfel, Philippe
               and Leroy, Vincent and Rogez, Gregory},
  booktitle = {IEEE/CVF International Conference on Computer Vision (ICCV)},
  year      = {2025},
}

@inproceedings{gvhmr,
  title     = {World-Grounded Human Motion Recovery via Gravity-View Coordinates},
  author    = {Shen, Zehong and Pi, Huaijin and Xia, Yan and Cen, Zhi and Peng, Sida and Hu, Zechen
               and Bao, Hujun and Hu, Ruizhen and Zhou, Xiaowei},
  booktitle = {SIGGRAPH Asia},
  year      = {2024}
}

@misc{watch,
  title  = {{WATCH}: World-aware Allied Trajectory and Pose Reconstruction for Camera and Human},
  author = {Ying, Qijun and Hu, Zhongyuan and Zhang, Rui and Li, Ronghui and Lu, Yu
            and Zeng, Zijiao},
  year   = {2025},
  eprint = {2509.04600},
  archivePrefix = {arXiv}
}

@inproceedings{josh,
  title     = {Joint Optimization for {4D} Human-Scene Reconstruction in the Wild},
  author    = {Liu, Zhizheng and Lin, Joe and Wu, Wayne and Zhou, Bolei},
  booktitle = {International Conference on Learning Representations (ICLR)},
  year      = {2026}
}

@inproceedings{prompthmr,
  title     = {{PromptHMR}: Promptable Human Mesh Recovery},
  author    = {Wang, Yufu and Sun, Yu and Patel, Priyanka and Daniilidis, Kostas
               and Black, Michael J. and Kocabas, Muhammed},
  booktitle = {IEEE/CVF Conference on Computer Vision and Pattern Recognition (CVPR)},
  year      = {2025},
  note      = {arXiv:2504.06397; video variant numbers as reported in~\cite{watch}}
}

@inproceedings{trace,
  title     = {{TRACE}: 5D Temporal Regression of Avatars with Dynamic Cameras in 3D Environments},
  author    = {Sun, Yu and Bao, Qian and Liu, Wu and Mei, Tao and Black, Michael J.},
  booktitle = {CVPR},
  year      = {2023}
}

@inproceedings{hmr2,
  title     = {Humans in 4{D}: Reconstructing and Tracking Humans with Transformers},
  author    = {Goel, Shubham and Pavlakos, Georgios and Rajasegaran, Jathushan and Kanazawa, Angjoo and Malik, Jitendra},
  booktitle = {IEEE/CVF International Conference on Computer Vision (ICCV)},
  year      = {2023}
}

@article{vitposepp,
  title     = {{ViTPose++}: Vision Transformer for Generic Body Pose Estimation},
  author    = {Xu, Yufei and Zhang, Jing and Zhang, Qiming and Tao, Dacheng},
  journal   = {IEEE Transactions on Pattern Analysis and Machine Intelligence},
  volume    = {46},
  number    = {2},
  pages     = {1212--1230},
  year      = {2024}
}

@inproceedings{3dpw,
  title     = {Recovering Accurate 3D Human Pose in the Wild Using {IMU}s and a Moving Camera},
  author    = {von Marcard, Timo and Henschel, Roberto and Black, Michael J. and Rosenhahn, Bodo and Pons-Moll, Gerard},
  booktitle = {European Conference on Computer Vision (ECCV)},
  pages     = {601--617},
  year      = {2018}
}

@inproceedings{hsfm,
  title     = {Reconstructing People, Places, and Cameras},
  author    = {M{\"u}ller, Lea and Choi, Hongsuk and Zhang, Anthony and Yi, Brent
               and Malik, Jitendra and Kanazawa, Angjoo},
  booktitle = {IEEE/CVF Conference on Computer Vision and Pattern Recognition (CVPR)},
  year      = {2025},
}

@inproceedings{rich,
  title     = {Capturing and Inferring Dense Full-Body Human-Scene Contact},
  author    = {Huang, Chun-Hao P. and Yi, Hongwei and H{\"o}schle, Markus and Safroshkin, Matvey
               and Alexiadis, Tsvetelina and Polikovsky, Senya and Scharstein, Daniel
               and Black, Michael J.},
  booktitle = {IEEE/CVF Conference on Computer Vision and Pattern Recognition (CVPR)},
  year      = {2022}
}

@inproceedings{smplx,
  title     = {Expressive Body Capture: {3D} Hands, Face, and Body from a Single Image},
  author    = {Pavlakos, Georgios and Choutas, Vasileios and Ghorbani, Nima and Bolkart, Timo and Osman, Ahmed A. A. and Tzionas, Dimitrios and Black, Michael J.},
  booktitle = {CVPR},
  year      = {2019}
}

@inproceedings{hac,
  title     = {Humans as Checkerboards: Calibrating Camera Motion Scale for World-Coordinate
               Human Mesh Recovery},
  author    = {Yang, Fengyuan and Gu, Kerui and Nguyen, Ha Linh and Tse, Tze Ho Elden and Yao, Angela},
  booktitle = {IEEE/CVF International Conference on Computer Vision (ICCV)},
  year      = {2025},
}

@inproceedings{sceneprobes,
  title     = {People as Scene Probes},
  author    = {Wang, Yifan and Curless, Brian and Seitz, Steve},
  booktitle = {European Conference on Computer Vision (ECCV)},
  year      = {2020},
}

@article{metrichmsr,
  title   = {MetricHMSR: Metric Human Mesh and Scene Recovery from Monocular Images},
  author  = {Song, Chentao and Zhang, He and Yuan, Haolei and Lin, Haozhe and Tao, Jianhua
             and Zhang, Hongwen and Yu, Tao},
  journal = {arXiv preprint arXiv:2506.09919},
  year    = {2025},
}

@inproceedings{mapanything,
  title     = {MapAnything: Universal Feed-Forward Metric 3{D} Reconstruction},
  author    = {Keetha, Nikhil and M{\"u}ller, Norman and Sch{\"o}nberger, Johannes and Porzi, Lorenzo
               and Zhang, Yuchen and Fischer, Tobias and Knapitsch, Arno and Zauss, Duncan
               and Weber, Ethan and Antunes, Nelson and Luiten, Jonathon and Lopez-Antequera, Manuel
               and Rota Bul{\`o}, Samuel and Richardt, Christian and Ramanan, Deva
               and Scherer, Sebastian and Kontschieder, Peter},
  booktitle = {International Conference on 3D Vision (3DV)},
  year      = {2026},
}

@inproceedings{metricanything,
  title     = {MetricAnything: Scaling Metric Depth Pretraining with Noisy Heterogeneous Sources},
  author    = {Ma, Baorui and Yang, Jiahui and Di, Donglin and Zhang, Xuancheng and Cui, Jianxun
               and Li, Hao and Xie, Yan and Chen, Wei},
  booktitle = {European Conference on Computer Vision (ECCV)},
  year      = {2026},
}

@article{gush3r,
  title   = {GUSH3R: Everyone Everywhere All at Once as Gaussians},
  author  = {Abe, Keito and Shiohara, Kaede and Otonari, Takashi and Yamasaki, Toshihiko},
  journal = {arXiv preprint arXiv:2607.05243},
  year    = {2026},
}

@inproceedings{noisystudent,
  title     = {Self-Training With Noisy Student Improves {ImageNet} Classification},
  author    = {Xie, Qizhe and Luong, Minh-Thang and Hovy, Eduard and Le, Quoc V.},
  booktitle = {IEEE/CVF Conference on Computer Vision and Pattern Recognition (CVPR)},
  year      = {2020},
}

@inproceedings{pseudolabel,
  title     = {Pseudo-Label: The Simple and Efficient Semi-Supervised Learning Method for Deep Neural Networks},
  author    = {Lee, Dong-Hyun},
  booktitle = {ICML Workshop on Challenges in Representation Learning},
  year      = {2013},
}

@inproceedings{dust3r,
  title     = {{DUSt3R}: Geometric 3{D} Vision Made Easy},
  author    = {Wang, Shuzhe and Leroy, Vincent and Cabon, Yohann and Chidlovskii, Boris and Revaud, J{\'e}r{\^o}me},
  booktitle = {IEEE/CVF Conference on Computer Vision and Pattern Recognition (CVPR)},
  year      = {2024},
}

@inproceedings{monst3r,
  title     = {{MonST3R}: A Simple Approach for Estimating Geometry in the Presence of Motion},
  author    = {Zhang, Junyi and Herrmann, Charles and Hur, Junhwa and Jampani, Varun and Darrell, Trevor
               and Cole, Forrester and Sun, Deqing and Yang, Ming-Hsuan},
  booktitle = {International Conference on Learning Representations (ICLR)},
  year      = {2025},
}

@inproceedings{easi3r,
  title     = {{Easi3R}: Estimating Disentangled Motion from {DUSt3R} Without Training},
  author    = {Chen, Xingyu and Chen, Yue and Xiu, Yuliang and Geiger, Andreas and Chen, Anpei},
  booktitle = {IEEE/CVF International Conference on Computer Vision (ICCV)},
  year      = {2025},
}

@inproceedings{probe3d,
  title     = {Probing the 3{D} Awareness of Visual Foundation Models},
  author    = {El Banani, Mohamed and Raj, Amit and Maninis, Kevis-Kokitsi and Kar, Abhishek and Li, Yuanzhen
               and Rubinstein, Michael and Sun, Deqing and Guibas, Leonidas and Johnson, Justin and Jampani, Varun},
  booktitle = {IEEE/CVF Conference on Computer Vision and Pattern Recognition (CVPR)},
  year      = {2024},
}

@inproceedings{dino,
  title     = {Emerging Properties in Self-Supervised Vision Transformers},
  author    = {Caron, Mathilde and Touvron, Hugo and Misra, Ishan and J{\'e}gou, Herv{\'e} and Mairal, Julien
               and Bojanowski, Piotr and Joulin, Armand},
  booktitle = {IEEE/CVF International Conference on Computer Vision (ICCV)},
  year      = {2021},
}

@inproceedings{linearprobes,
  title     = {Understanding Intermediate Layers Using Linear Classifier Probes},
  author    = {Alain, Guillaume and Bengio, Yoshua},
  booktitle = {International Conference on Learning Representations (ICLR), Workshop Track},
  year      = {2017},
}

@inproceedings{monodepth2,
  title     = {Digging Into Self-Supervised Monocular Depth Estimation},
  author    = {Godard, Cl{\'e}ment and Mac Aodha, Oisin and Firman, Michael and Brostow, Gabriel J.},
  booktitle = {IEEE/CVF International Conference on Computer Vision (ICCV)},
  year      = {2019},
}

@inproceedings{eigen,
  title     = {Depth Map Prediction from a Single Image Using a Multi-Scale Deep Network},
  author    = {Eigen, David and Puhrsch, Christian and Fergus, Rob},
  booktitle = {Advances in Neural Information Processing Systems (NeurIPS)},
  year      = {2014},
}

@inproceedings{coin,
  title     = {{COIN}: Control-Inpainting Diffusion Prior for Human and Camera Motion Estimation},
  author    = {Li, Jiefeng and Yuan, Ye and Rempe, Davis and Zhang, Haotian and Molchanov, Pavlo and Lu, Cewu
               and Kautz, Jan and Iqbal, Umar},
  booktitle = {European Conference on Computer Vision (ECCV)},
  year      = {2024},
}

@inproceedings{lora,
  title     = {{LoRA}: Low-Rank Adaptation of Large Language Models},
  author    = {Hu, Edward J. and Shen, Yelong and Wallis, Phillip and Allen-Zhu, Zeyuan and Li, Yuanzhi
               and Wang, Shean and Wang, Lu and Chen, Weizhu},
  booktitle = {International Conference on Learning Representations (ICLR)},
  year      = {2022},
}

@article{vggtomega,
  title   = {{VGGT}-$\Omega$},
  author  = {Wang, Jianyuan and Chen, Minghao and Zhang, Shangzhan and Karaev, Nikita and Sch{\"o}nberger, Johannes
             and Labatut, Patrick and Bojanowski, Piotr and Novotny, David and Vedaldi, Andrea and Rupprecht, Christian},
  journal = {arXiv preprint arXiv:2605.15195},
  year    = {2026},
}

@inproceedings{amb3r,
  title     = {{AMB3R}: Accurate Feed-forward Metric-scale 3D Reconstruction with Backend},
  author    = {Wang, Hengyi and Agapito, Lourdes},
  booktitle = {IEEE/CVF Conference on Computer Vision and Pattern Recognition (CVPR)},
  year      = {2026},
}
\bibliographystyle{iclr2027_conference}

\clearpage
\appendix
\raggedbottom
\begin{center}
 {\large\bfseries Supplementary Material}\\[0.2em]
 {\normalsize WildHSR: Metric Feed-Forward 4D People-Scene Reconstruction from a 3D Foundation Model}
\end{center}
\vspace{-0.4em}
\noindent\currentrevision{Appendices A--E cover the association protocol, runtime, RICH evaluation, scale calibration, and scene reconstruction. Appendix F presents qualitative results. Appendices G--K cover ablations, people-scene consistency, identity and pelvis probes, robustness, and failure modes.}\par
\FloatBarrier
\section{Temporal Association Protocol}
\label{app:association}
Table~\ref{tab:identity_assoc} reports a controlled component experiment, not the final end-to-end pipeline. WildHSR variants use fixed Multi-HMR proposals and frozen bodies, camera-frame pelvis positions, non-overlapping windows, and an identity projection trained on 3DPW only. Assignment thresholds were tuned on four validation sequences. Human3R retains its native pipeline; the comparison therefore tests tracking behavior, not identical upstream reconstructions.

\FloatBarrier
\section{Computational Cost}
\label{app:compute}
\textbf{Training configuration.} The released VGGT-$\Omega$ and Multi-HMR base weights remain frozen. The backbone
LoRA uses rank 16, $\alpha=32$ and dropout $0.05$ on the linear layers of the last four frame-attention and last
four global-attention blocks (\vggtLoraParams{} parameters); the mesh adapter has \loraParams{} parameters. Every
Scale Readout in Table~\ref{tab:scale_stages} uses a 512-dimensional query, four decoder layers, eight attention
heads, batch size 16 and 6,000 pretraining steps. It is initialized on BEDLAM2, pretrained from ruler labels with
the backbone frozen, then jointly fine-tuned with the backbone LoRA on the standard RICH and 3DPW training splits
using AdamW at $10^{-4}$. Standard validation partitions are used for model selection. \revision{The Pelvis Readout
uses each proposal's HMR tokens to produce the image-space hip-midpoint location and query. The location is trained
against the projected ground-truth SMPL-X pelvis; decoded-body losses train the query and fusion, without an
attention-map target.} The identity readout pools
layer-13 query and key vectors around each \revision{Pelvis Readout location}, maps them to a 128-dimensional unit vector with a
two-layer MLP, and is trained with supervised contrastive and dustbin-aware soft-assignment losses on the BEDLAM2,
RICH-train and 3DPW-train identities while both base networks remain frozen. Training windows contain 17 frames for
BEDLAM2 and 32 frames for real video. \currentrevision{These are training clip lengths, not the inference window:
the reported evaluations use 100-frame windows with stride 50. No window-length ablation is reported.}

\begin{table}[H]
\centering
\small
\caption{\textbf{Compute on one GPU, same 60-frame clip.} \currentrevision{End-to-end throughput and peak memory; FLOPs count traced model stages.}}
\label{tab:compute}
\setlength{\tabcolsep}{5pt}
\begin{tabular}{lccccc}
\toprule
method & params run$\downarrow$ & TFLOPs/frame$\downarrow$ & fps$\uparrow$ & s / 60 frames$\downarrow$ & peak GPU$\downarrow$ \\
\midrule
Human3R & \bestc{1.17B} & n/a & \thirdc{8.8} & \thirdc{6.8} & \bestc{5.95\,GB} \\
UniSH & 1.86B & \thirdc{7.29} & 4.9 & 12.2 & \secondc{7.06\,GB} \\
JOSH3R & \secondc{1.50B} & \bestc{3.77} & 6.1 & 9.8 & \thirdc{7.55\,GB} \\
\midrule
\name{} ($512\times288$) & \thirdc{1.52B} & \secondc{6.94} & \bestc{12.6} & \bestc{4.8} & 8.85\,GB \\
\name{} ($688\times384$, ours) & \thirdc{1.52B} & 11.43 & \secondc{10.1} & \secondc{6.0} & 11.22\,GB \\
\bottomrule
\end{tabular}
\end{table}

\FloatBarrier
\currentrevision{Table}~\ref{tab:compute} \currentrevision{reports throughput, peak GPU memory and traced model FLOPs on the first 60 frames of EMDB-2 \texttt{24\_outdoor\_long\_walk}. We use one} \benchGPU{} \currentrevision{per method, two warmup passes and three timed passes, excluding model loading. Each system uses its evaluated resolution. FPS measures the full pipeline; FLOPs cover traced neural stages only. The parameter count includes every loaded model in the inference path.}

\currentrevision{At its main} $688\times384$ \currentrevision{resolution,} \name{} \currentrevision{processes} $10.1$ \currentrevision{frames per second, the highest end-to-end throughput among the configurations in Table}~\ref{tab:compute}\currentrevision{. Its body encoder accounts for} $62\%$ \currentrevision{of runtime, making that stage the clearest target for further speedups.}

\FloatBarrier
\section{RICH Evaluation Protocol}
\label{app:rich}
\textbf{Identifying the target subject.} RICH annotates one subject while up to \richMaxPeople{}
people are visible, so every method must decide which reconstruction to score; \richCrowdSegs{}
of our \richSegs{} segments contain more than one person. Motion and centre confidence alone do not reliably separate a
smoothly walking, confidently detected bystander from the annotated subject. Our association instead matches the
projected VGGT identity descriptor against the track memory and uses metric pelvis motion only as a complementary
cue. A dustbin permits the annotated person to be temporarily missing rather than forcing a match to a bystander.
No image-centre prior or capture-rig-specific rule enters the method.

\textbf{Scale-error metric.} Throughout the paper, ``median multiplicative scale error (\%)'' denotes
\begin{equation}
E_{\sigma}=100\left[
\exp\left(\operatorname*{median}_{i}
\left|\log\frac{\sigma^{\mathrm{pred}}_{i}}{\sigma^{\mathrm{gt}}_{i}}\right|\right)-1
\right].
\label{eq:scale-error}
\end{equation}
This is symmetric in log scale before conversion to a percentage: predicting either twice or half the
ground-truth scale gives $E_{\sigma}=100\%$.

\currentrevision{We use the published RICH test split and evaluator, though training may differ. Median segment displacement is} $0.52$\,m\currentrevision{, so RTE magnifies small trajectory errors.}

\FloatBarrier
\section{Held-Out Ruler Calibration on RICH}
\label{app:ruler_calibration}
\noindent\begin{minipage}[t]{0.47\textwidth}
\vspace{0pt}
\centering
\small
\setlength{\abovecaptionskip}{0pt}
\captionof{table}{\textbf{Ruler calibration on held-out RICH.} Median multiplicative scale error (\%).}
\label{tab:ruler_rich}
\setlength{\tabcolsep}{3pt}
\begin{tabular}{lc}
\toprule
scale source & error$\downarrow$ \\
\midrule
body-ruler pseudo-label & \thirdc{14.1} \\
Scale Readout, ruler-pretrained & \secondc{8.4} \\
Scale Readout, metric fine-tuned & \bestc{7.2} \\
\bottomrule
\end{tabular}
\end{minipage}\hfill
\begin{minipage}[t]{0.49\textwidth}
\vspace{0pt}
\centering\small
\setlength{\abovecaptionskip}{0pt}
\captionof{table}{\textbf{7Scenes scene accuracy.} Mean point-map accuracy in cm ($\downarrow$);}
\label{tab:scene_metric}
\setlength{\tabcolsep}{5pt}
\begin{tabular}{lc}
\toprule
method & Acc. (cm)$\downarrow$ \\
\midrule
Spann3R & 4.81 \\
MapAnything & 3.48 \\
CUT3R & 2.88 \\
VGGT & \thirdc{2.32} \\
AMB3R & \secondc{1.75} \\
\textbf{\name{}} & \bestc{1.66} \\
\cmidrule(lr){1-2}
\scriptsize $\Delta$ vs best & \scriptsize \sceneVsBestAcc{} \\
\bottomrule
\end{tabular}
\end{minipage}

The body ruler can inherit systematic error from the metric body teacher. Table~\ref{tab:ruler_rich}
compares the scale pathway's stages on the same held-out RICH examples using the median multiplicative
scale error of Eq.~\eqref{eq:scale-error}. Error falls from $14.1\%$ for the pseudo-label to $8.4\%$
after ruler pretraining and $7.2\%$ after exact metric fine-tuning.

\FloatBarrier
\section{Person-Free Scene Reconstruction}
\label{app:scene_metric}
\currentrevision{Table}~\ref{tab:scene_metric} \currentrevision{compares published 7Scenes point-map accuracy under the same scale-aligned protocol}~\citep{amb3r}. \name{} \currentrevision{has the lowest error at} $1.66$\,cm \currentrevision{versus} $1.75$\,cm \currentrevision{for AMB3R. This tests scene geometry, not metric scale without visible people.}

\FloatBarrier
\section{Qualitative Results}
\label{app:qual}
Figures~\ref{fig:qual}--\ref{fig:itw} show metric people, cameras and scene geometry across varied motion, longer videos and crowds. Rigid camera-path alignment preserves predicted scale in the people-scene views; similarity alignment isolates trajectory shape and drift.

\noindent\begin{minipage}{\textwidth}
{\centering\includegraphics[width=0.92\linewidth]{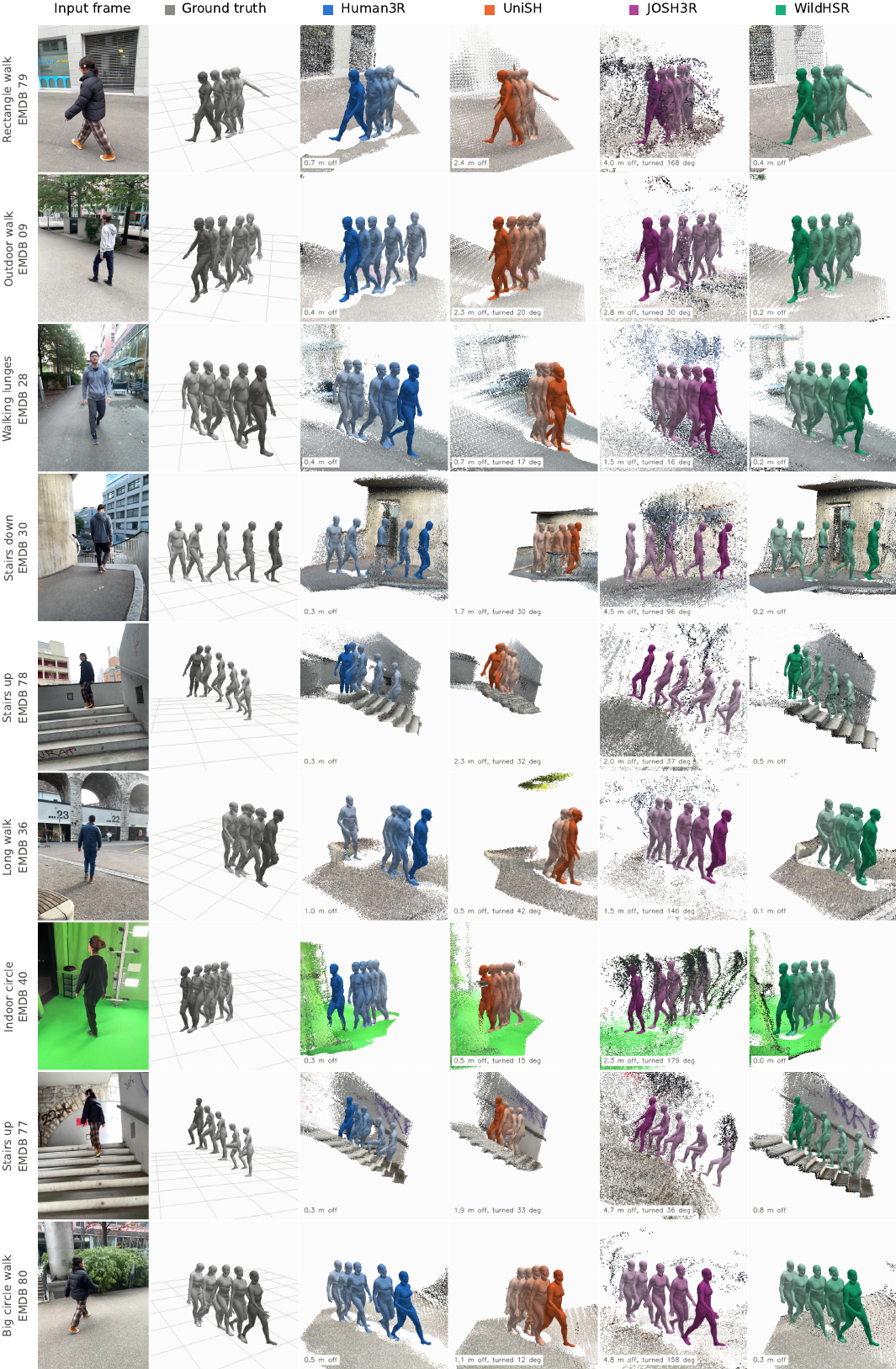}\par}\vspace{7pt}
\captionof{figure}{\textbf{Metric people-scene placement} on nine EMDB-2 clips. Each method's body history and scene retain predicted scale after rigid camera-path alignment. Labels report pelvis offset in metres; UniSH and JOSH3R also show the view rotation required for alignment.}
\label{fig:qual}

\textbf{What the alignment shows.} Across walks, stairs and lunges, \name{} has the lowest pelvis offset on seven of nine clips. Human3R is closer on two stairs-up clips ($0.3$ versus $0.5$ and $0.8$\,m). The body history stays positioned relative to its reconstructed ground, linking human placement to a coherent metric people-scene reconstruction throughout each motion.
\end{minipage}

\par\medskip\noindent\begin{minipage}[t]{0.48\textwidth}
\vspace{0pt}
{\centering\includegraphics[width=\linewidth]{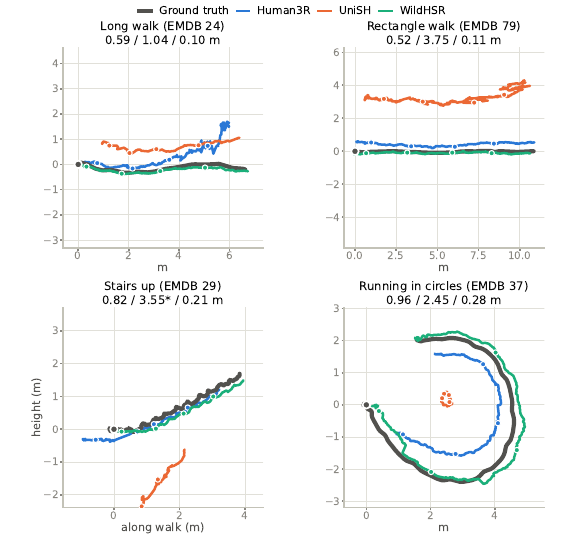}\par}
\end{minipage}\hfill
\begin{minipage}[t]{0.49\textwidth}
\vspace{0pt}
\captionof{figure}{\textbf{Global human trajectories} on four EMDB-2 clips after WA-MPJPE alignment. Stairs plot height; other panels are top-down. Numbers give mean path error (Human3R / UniSH / \name{}); dots mark 2\,s intervals and $*$ a path leaving the plot.}
\label{fig:traj}
\textbf{What the paths show.} \currentrevision{Figure}~\ref{fig:traj} \currentrevision{tests whether placement stays coherent through long walks, turns and stairs.} \name{} \currentrevision{follows the route and ascent, while Human3R drifts on the long walk and UniSH loses the turning path. Mean path error is} $0.10$ \currentrevision{to} $0.28$\,m \currentrevision{for} \name{}, $0.52$ \currentrevision{to} $0.96$\,m \currentrevision{for Human3R and} $1.04$ \currentrevision{to} $3.75$\,m \currentrevision{for UniSH. These paths are similarity-aligned; Fig.}~\ref{fig:map} \currentrevision{separately tests metric scale.}
\end{minipage}

\par\smallskip\noindent\begin{minipage}{\textwidth}
{\centering\includegraphics[width=1\linewidth]{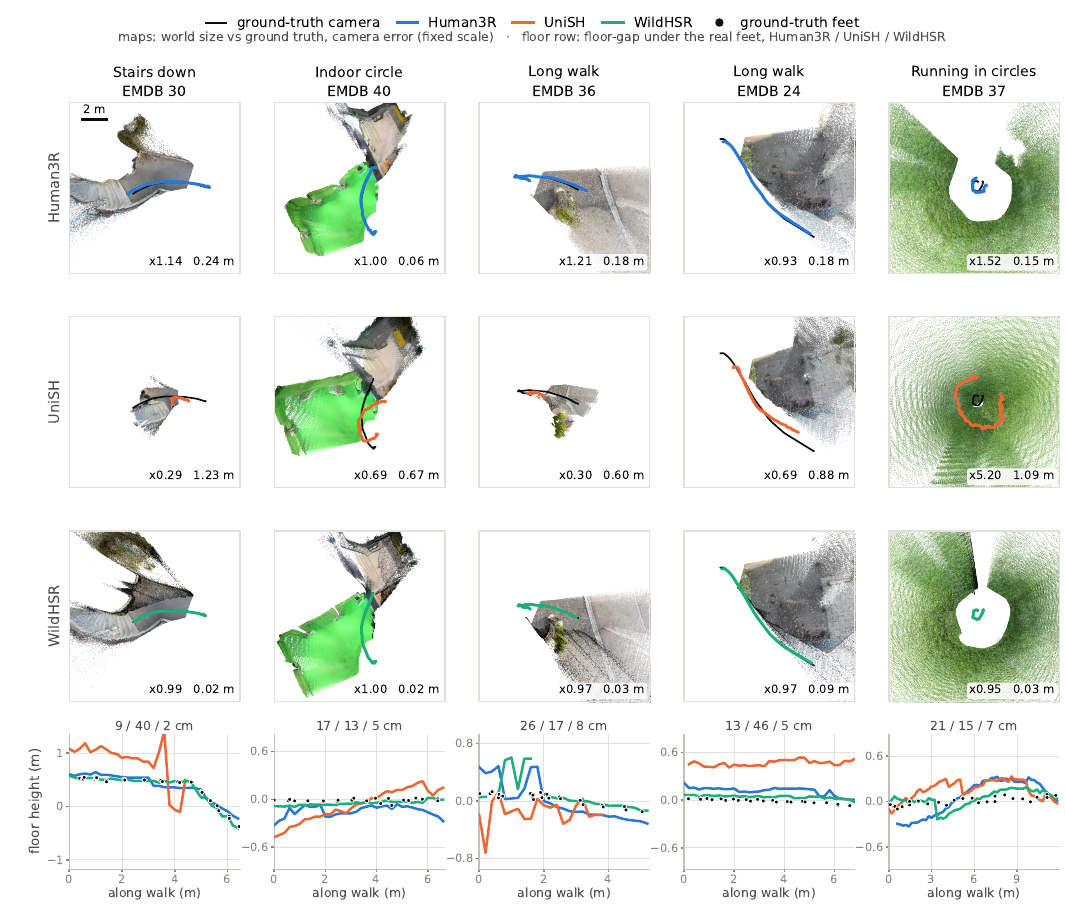}\par}\vspace{7pt}
\captionof{figure}{\textbf{Metric scene, camera and ground.} Top three rows: person-free scene and camera path after rigid alignment, preserving predicted scale; labels give world-size ratio and mean camera error. Bottom: reconstructed floor height along the walk against the ground-truth feet; values are median foot-floor gap for Human3R / UniSH / \name{}.}
\label{fig:map}
\textbf{What the scene shows.} \name{} reconstructs all five worlds at $0.95$ to $1.00\times$ ground-truth size with $0.02$ to $0.09$\,m mean camera error. Its median foot-floor gap is $2$ to $8$\,cm, compared with $9$ to $26$\,cm for Human3R and $4$ to $46$\,cm for UniSH. The scene scale, camera path and ground beneath the person remain mutually consistent, giving the human trajectory a metric reference in the reconstructed environment.
\end{minipage}

\par\medskip\noindent\begin{minipage}{\textwidth}
{\centering\includegraphics[width=\linewidth]{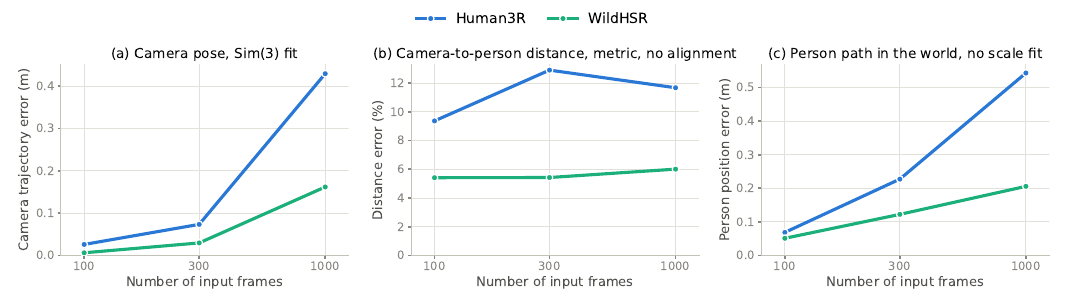}\par}\vspace{7pt}
\captionof{figure}{\textbf{Accuracy against video length}, averaged over six EMDB-2 clips of at least 1000 frames. (a) Camera trajectory error after Sim(3) alignment. (b) Metric camera-to-person distance error without alignment. (c) World-frame person-path error after rigid alignment, retaining scale error.}
\label{fig:length}
\end{minipage}

\par\medskip\noindent\begin{minipage}{\textwidth}
{\centering\includegraphics[width=0.95\linewidth]{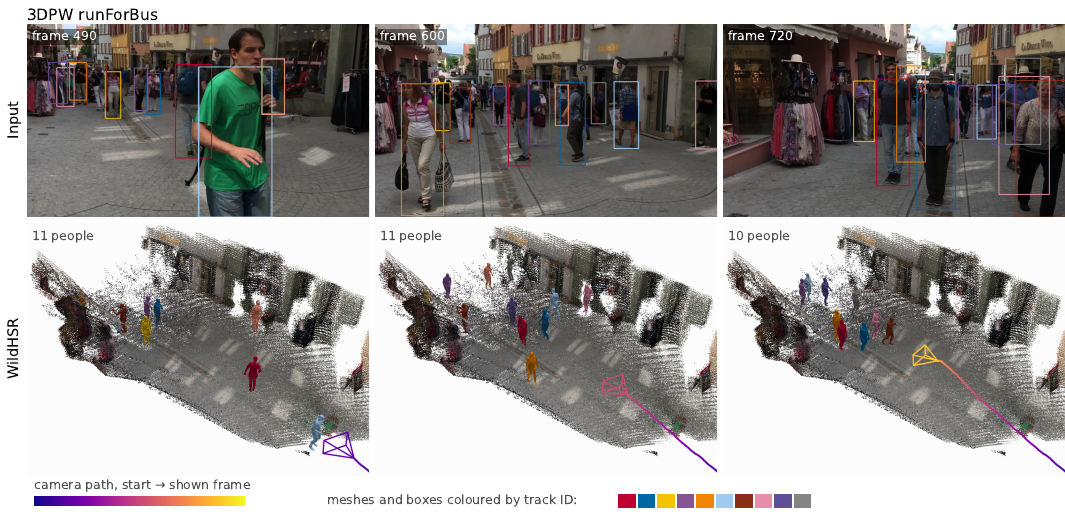}\par}\vspace{7pt}
\captionof{figure}{\textbf{Crowded in-the-wild reconstruction} on 3DPW \texttt{downtown\_runForBus}. Top: input and persistent tracks. Bottom: metric scene, camera trajectory and bodies colored by identity. The sequence contains 14 tracks over 10\,s, with 10 to 11 people reconstructed per frame.}
\label{fig:itw}
\textbf{What the crowd shows.} \name{} maintains 14 tracks over 10\,s and reconstructs 10 to 11 people per frame in one scene. For the two people with 3DPW annotations, hip-depth error is $0.29$\,m for \name{} and $1.20$\,m for Human3R, while local pose is similar (PA-MPJPE $55.3$ versus $51.7$\,mm). The larger improvement in placement shows the value of relating each person's body to the shared scene and camera estimate.

\textbf{What longer videos show.} From 100 to 1000 frames in Fig.~\ref{fig:length}, \name{} accumulates less camera and human-path error than Human3R, while its metric camera-to-person distance error stays near $5$ to $6\%$. The method maintains person placement in its reconstructed world as the observed path grows. \name{} processes each complete prefix; Human3R runs causally.
\end{minipage}

\FloatBarrier
\section{Additional Ablations}
\label{app:ablations}
\noindent\begin{minipage}[t]{0.49\textwidth}
\vspace{0pt}
\centering
\setlength{\abovecaptionskip}{0pt}
\captionof{table}{\textbf{Ablations on EMDB-2.} Each row substitutes one component of the full system.}
\label{tab:ablations}
\scriptsize
\setlength{\tabcolsep}{2pt}
\begin{tabular}{lccc}
\toprule
 & \multicolumn{3}{c}{EMDB-2} \\
\cmidrule(lr){2-4}
configuration & WA$\downarrow$ & W$\downarrow$ & RTE$\downarrow$ \\
\midrule
\textbf{\name{} (full)} & \bestc{\oursWA{}} & \bestc{\oursW{}} & \bestc{\oursRTE{}} \\
\midrule
synthetic-only scale & \thirdc{\ablNoRulerWA{}} & \thirdc{\ablNoRulerW{}} & \thirdc{\ablNoRulerRTE{}} \\
$-$ scene contact & \secondc{\ablNoContactWA{}} & \secondc{\ablNoContactW{}} & \secondc{\ablNoContactRTE{}} \\
\bottomrule
\end{tabular}
\end{minipage}\hfill
\begin{minipage}[t]{0.47\textwidth}
\vspace{0pt}
\textbf{Scale pathway.} Table~\ref{tab:ablations} shows the largest loss when the final two-stage Scale Readout is replaced by the best regressor trained on synthetic metric ground truth: $\noRulerWARatio\times$ WA and $13\times$ RTE. Because the final head combines ruler pretraining with ground-truth fine-tuning, this substitution does not isolate their separate contributions.
\end{minipage}
\par\medskip
\FloatBarrier

\section{Plausibility and Metric Scale on EMDB-2}
\label{app:plausibility}
\noindent Table~\ref{tab:contact_scale} measures people-scene consistency on all 25 EMDB-2 sequences. Foot gap compares the sole with each method's reconstructed floor, and sliding measures toe speed during ground-truth contact. World size compares camera-trajectory scale with ground truth (ideal: one), while camera error aligns rigidly without rescaling. \name{} reduces the foot gap to 3.6\,cm versus 12.7\,cm for Human3R and 15.5\,cm for UniSH. Its 6\% world-size error and 0.13\,m camera error support placing bodies and scene in a shared metric frame; contact is a training term, not an inference correction.

\begin{table}[H]
\centering
\small
\caption{\textbf{People-scene consistency on EMDB-2.} Five diagnostics use the first 300 frames of each sequence; WA-MPJPE uses full sequences. Shades rank methods per column, with world size ranked by distance from one.}
\label{tab:contact_scale}
\setlength{\tabcolsep}{3pt}
\begin{tabular}{lcccccc}
\toprule
method & foot gap (cm)$\downarrow$ & sliding (cm/s)$\downarrow$ & world size ($\times$GT) & size err.$\downarrow$ & camera (m)$\downarrow$ & WA (mm)$\downarrow$ \\
\midrule
Human3R & \secondc{12.7} & \secondc{22.6} & \secondc{1.03} & \secondc{15\%} & \secondc{0.20} & \secondc{\hthreeWA{}} \\
UniSH & \thirdc{15.5} & \thirdc{28.1} & \thirdc{0.47} & \thirdc{73\%} & \thirdc{0.98} & \thirdc{\unishWA{}} \\
\textbf{WildHSR} & \bestc{3.6} & \bestc{17.4} & \bestc{0.99} & \bestc{6\%} & \bestc{0.13} & \bestc{66.3} \\
\bottomrule
\end{tabular}
\end{table}

\noindent\begin{minipage}[t]{0.47\textwidth}
\vspace{0pt}
\centering
\includegraphics[width=\linewidth, trim=7pt 13pt 229pt 27pt, clip]{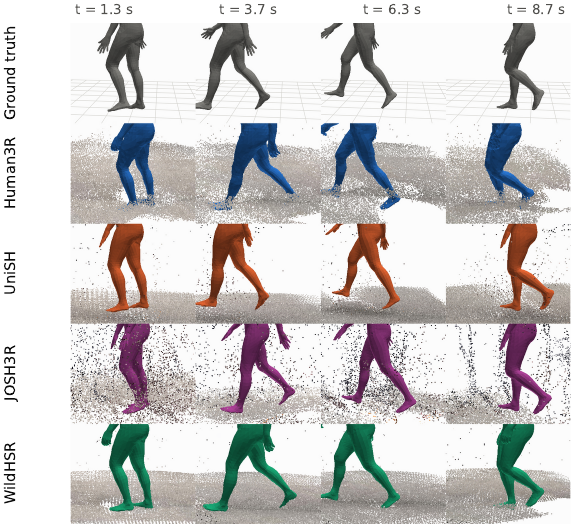}
\setlength{\abovecaptionskip}{3pt}
\captionof{figure}{\textbf{Feet on the reconstructed ground} (EMDB 79), each method on its own scene, at four moments with the same viewing direction and zoom.}
\label{fig:foot}
\end{minipage}\hfill
\begin{minipage}[t]{0.49\textwidth}
\vspace{0pt}
Fig.~\ref{fig:foot} shows feet relative to each method's reconstructed ground; Table~\ref{tab:contact_scale} extends the measurement to every EMDB-2 sequence.

\textbf{How Table~\ref{tab:contact_scale} is measured.} Feet-to-ground and sliding are read at each method's own metric scale: the method is
placed in the ground-truth world by a rotation and translation only alignment of its camera trajectory, so no
ground-truth scale is imposed on its body or on its scene. \emph{Feet-to-ground} is, per sequence, the median over frames
of the absolute height gap between the lower foot sole and the floor of the method's \emph{own} reconstructed scene
directly beneath it, where that floor is the mode of scene-point heights within 25\,cm horizontally, gathered from frames
within $\pm$1\,s. EMDB provides no ground-truth scene, so the quantity measures how self-consistent a method's human and
scene are, not how accurate either one is.
\end{minipage}

\par\medskip
The fraction of frames carrying a floor estimate is $95\%$ for Human3R, $100\%$ for UniSH and $98\%$ for \name{};
admitting floors up to $2.5$\,m above the sole instead of $0.5$\,m moves the means to $12.7$, $16.9$ and $3.6$\,cm,
so the ordering is not an artefact of that threshold.

\textbf{Other diagnostics in Table~\ref{tab:contact_scale}.} \currentrevision{Sliding is median horizontal toe speed on contact frames. A ground-truth toe counts as contacting when its speed is below} $0.25$\,m/s \currentrevision{and its height is within} 5\,cm \currentrevision{of its local minimum, after five-frame smoothing. World size is the median ratio of reconstructed to ground-truth camera-trajectory scale (ideal: one). Size error averages} $|1-s|$ \currentrevision{over sequences, so a near-one median can still conceal large errors on individual sequences. Camera error is mean camera-centre distance after rotation and translation alignment only. WA-MPJPE repeats the full-benchmark values of Table}~\ref{tab:main}\currentrevision{, not a first-300-frame recomputation.}

\needspace{8\baselineskip}
\textbf{Physical plausibility.} Table~\ref{tab:ablations} shows that
\revision{contact regularization} improves WA while leaving W-MPJPE and RTE nearly unchanged. These metrics do not directly test body-ground consistency. Measured against the backbone's own reconstructed floor on all 25 EMDB-2 sequences, the
lowest body vertex floats \floatBefore{} above the floor without the term and
\floatAfter{} with it, with \penAfter{} of frames penetrating the floor by more than 5\,cm;
foot skate and vertex jitter are unchanged (\skateVal{}\,cm/frame, \jitterVal{}\,cm/frame$^2$).
These measurements complement the benchmark metrics by testing people-scene consistency directly.

\FloatBarrier
\section{Identity Features and Pelvis Localization}
\label{app:identity}
\textbf{Probe setup.} \currentrevision{We run the unmodified pretrained} VGGT-$\Omega$ \currentrevision{alone; the mesh model, fusion and adapters are absent. Person boxes and tracks select evaluation patches and provide identity labels, but are not inputs to the backbone.} VGGT-$\Omega$ \currentrevision{learns geometric matching on static scene points, with movable content excluded from that supervision}~\citep{vggtomega}. \currentrevision{The question is whether its matching representation also links moving people, even though no training target associates their identities.}

\textbf{Position control.} \currentrevision{A stationary person can be matched by image location alone, so the probe separately tests pairs where the person moves (box IoU below} $0.2$\currentrevision{). It compares the person's new position with the vacated position and other people. The aggregate result below, rather than a selected frame pair, motivates the intermediate-depth identity features.}

\begin{figure}[!t]
\centering
\includegraphics[width=\textwidth]{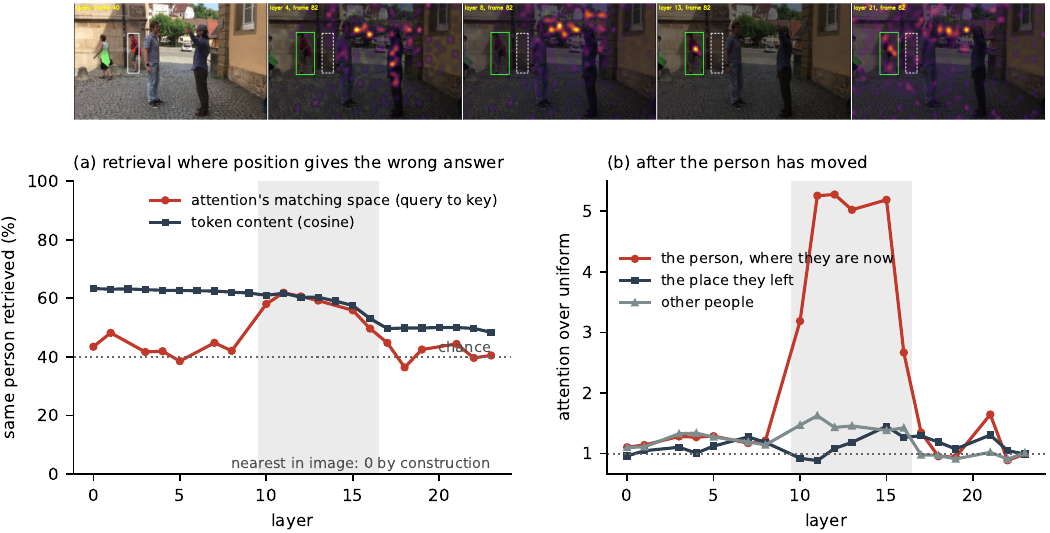}
\caption{\textbf{Position or person?} Pretrained VGGT-$\Omega$ by itself. Top: a query token on a walking person
(white box) and its global attention on a later frame, at layers 4, 8, 13 and 21; dashed box, where the person was;
green box, where they are. The example is the clip's median, not its best. Bottom, means over clips, layers 10 to 16
shaded: (a) same-person retrieval on pairs where the nearest candidate in the image is the wrong person, in the
attention's query and key space and in token content; (b) attention lift after the person has moved.}
\label{fig:tokens}
\end{figure}

\textbf{Retrieval over 1,500 clips.} For each 100-frame multi-person window, a patch on one person is matched to on-person patches in a later frame. A match is correct when its top candidate belongs to the same person. We evaluate output-token cosine similarity and pre-rotation query-key logits, including hard pairs where the nearest image-space candidate is the wrong person. Chance is $46\%$ overall and $40\%$ on hard pairs; moved-person pairs (box IoU below $0.2$) have $38\%$ chance. In query-key space, retrieval peaks at $80$ to $82\%$ in layers 11 to 15, including $70\%$ after motion and $60\%$ on hard pairs, then returns to chance in late layers. Figure~\ref{fig:tokens}(a) shows the intermediate-depth advantage on hard pairs, where matching by image location fails. Output tokens already retrieve at $85\%$ at layer 0, consistent with appearance information from the image encoder, but their retrieval decays with depth. This contrast motivates reading identity from intermediate query-key features rather than final tokens.

\textbf{After motion.} At layer 13, attention lift on the person's new position averages $5.0\times$ uniform, compared with $1.2\times$ on the vacated position and $1.5\times$ on other people among clips with sufficient moved pairs. The new position wins in $91\%$ of those clips, although overlap and small people weaken the effect. This control separates person correspondence from simply revisiting an image location; it does not by itself measure the trained tracker's accuracy.

\FloatBarrier

\textbf{What the probe does not show.} We have not run the same test on objects other than people, so we do not claim the
behaviour is specific to people: a model that matches any moving surface across frames would pass it, and for our
purpose that would serve equally. The probe alone also cannot establish that its correspondence improves tracking.
The association-component experiment in Table~\ref{tab:identity_assoc} tests how descriptor and motion cues
combine when upstream proposals and bodies are held fixed; it does not isolate their effect in the final
end-to-end pipeline.

\textbf{Identity readout and training.} The pelvis is the body root, giving a smoother motion anchor than distal joints and avoiding a head-to-root translation lever arm. Lower-body occlusion is examined in Appendix~\ref{app:occlusion}. For proposal $n$ in frame $t$, we take the QK-normalised layer-13 query and
key vectors before positional rotation, average each head over the $3\times3$ token neighbourhood around the pelvis
patch, concatenate all heads and both roles, and compute
\begin{equation}
 e_{t,n}=\operatorname{norm}_2\!\left(P\left([\bar q_{t,n};\bar k_{t,n}]\right)\right),
 \qquad P=\operatorname{Linear}\circ\operatorname{GELU}\circ\operatorname{Linear}\circ\operatorname{LayerNorm},
 \label{eq:identity_projection}
\end{equation}
where $e_{t,n}\in\mathbb{R}^{128}$. VGGT and the mesh branch remain frozen. We first train only $P$ with supervised
contrastive temperature $0.07$: observations of the same training identity at different times are positives and
co-occurring people are hard negatives. We then fine-tune $P$ through the soft assignment defined below using
negative log likelihood of the ground-truth match, birth and miss decisions. Identity groups are disjoint between
training and validation, and only BEDLAM2, RICH-train and 3DPW-train identities are used.

\textbf{State and dimensionless cost.} Association is applied after each window has been composed into the common
world frame of \S\ref{sec:stitch}. A live track $i$ stores a unit descriptor $m_i$, its last metric pelvis $p_i$, an
exponentially smoothed velocity $v_i$, last-seen time and track shape. At elapsed time $\Delta t$, its predicted pelvis
is $\hat p_i=p_i+v_i\Delta t$. Proposal $j$ has descriptor $e_j$, world pelvis $p_j$ and confidence $c_j$. We use
\begin{equation}
 \begin{split}
 d^{\mathrm{id}}_{ij}&=\tfrac12\left(1-\langle m_i,e_j\rangle\right),\\
 g_i&=0.5\,\mathrm{m}+\min(3\,\mathrm{m\,s^{-1}}\Delta t,3\,\mathrm{m}),\\
 d^{\mathrm{mot}}_{ij}&=\min\!\left(\lVert p_j-\hat p_i\rVert_2/g_i,1\right),\qquad
 d^{\mathrm{conf}}_j=1-c_j,\\
 C_{ij}&=d^{\mathrm{id}}_{ij}+d^{\mathrm{mot}}_{ij}+0.25d^{\mathrm{conf}}_j.
 \end{split}
 \label{eq:association_cost}
\end{equation}
The three component costs lie in $[0,1]$, while $C_{ij}\in[0,2.25]$. A real pair outside the motion gate,
$\lVert p_j-\hat p_i\rVert_2>g_i$, is invalid. These
weights and gates are fixed before benchmark evaluation.

\textbf{Births, misses and one-to-one hardening.} With $M$ tracks and $N$ proposals of confidence at least $0.5$, we
construct a square $(M+N)\times(N+M)$ cost matrix. Its top-left block is $C$. Track $i$ has one dedicated miss column
of cost $0.7$, proposal $j$ has one dedicated birth row of cost $0.7$, invalid dummy edges have infinite cost, and the
dummy-to-dummy block has zero cost. Log-domain Sinkhorn with temperature $0.1$ and 20 normalisation iterations produces
the soft matrix used by the assignment loss. At inference, Hungarian matching on its negative log probabilities gives
exactly one decision per real track and proposal. A matched real pair continues a track, a matched birth starts one,
and a matched miss preserves the track without fabricating an observation.

\textbf{Safe memory update.} A match updates memory only when its Sinkhorn probability is at least $0.7$:
$m_i\leftarrow\operatorname{norm}_2(0.9m_i+0.1e_j)$, while velocity uses the analogous $0.8/0.2$ update. Lower-confidence
matches keep the identity but do not change descriptor, velocity or shape memory. A missed track can be reactivated for
one second and is then terminated. Track shape is the median of confident matched observations only. Proposal selection,
Hungarian hardening, births, misses, memory updates and the shape median receive no gradient. Sinkhorn is differentiable
during training and analytic at inference; no test-time model fitting or backpropagation is used.

\FloatBarrier
\needspace{0.70\textheight}
\section{Robustness to Truncation and Occlusion}
\label{app:occlusion}
\begin{figure}[H]
{\centering\includegraphics[width=0.9\textwidth]{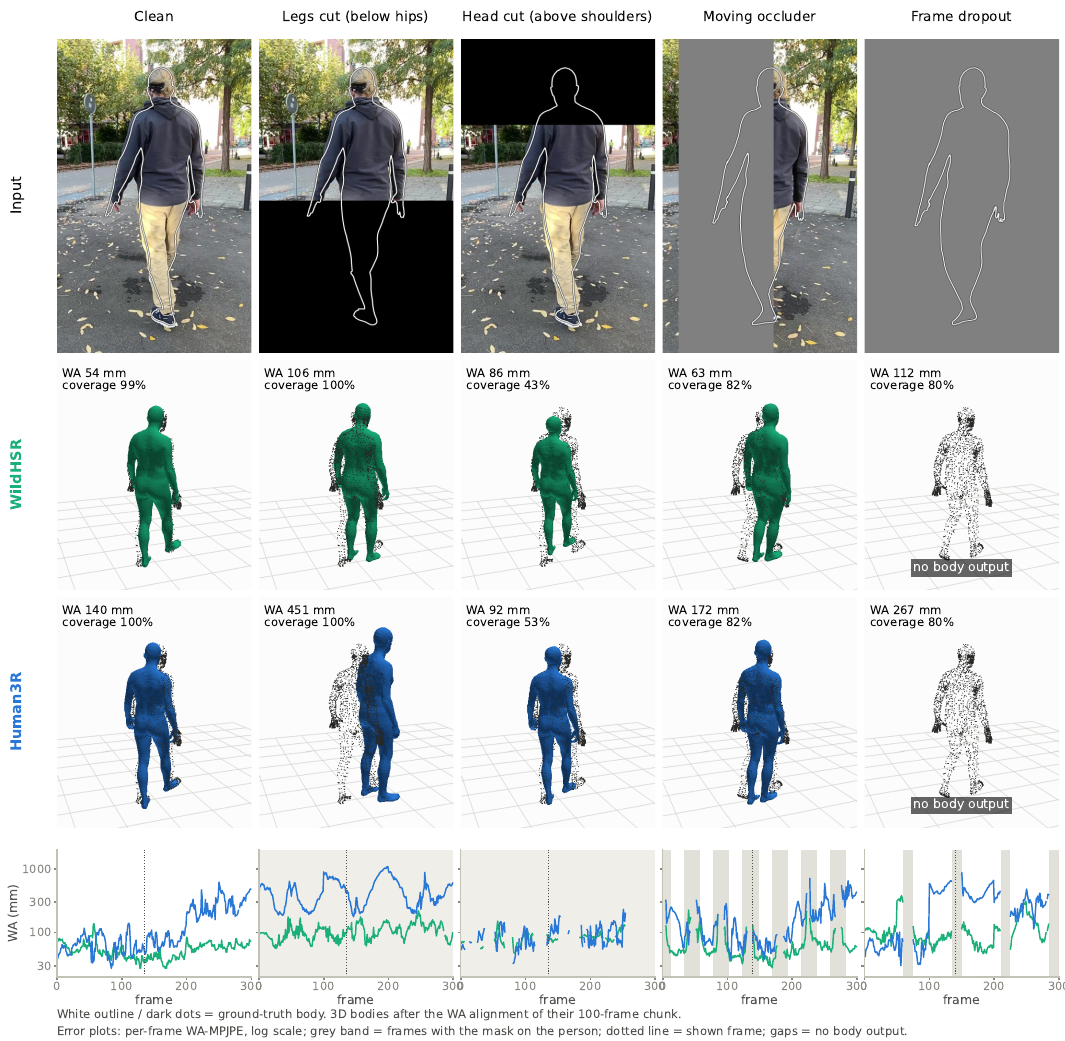}\par}\vspace{5pt}
\caption{\textbf{Truncation and occlusion} on a 300-frame EMDB-2 clip. Top: masked input and ground-truth outline. Middle: reconstructed bodies after WA alignment, with ground truth as dots. Bottom: per-frame WA-MPJPE on a log scale; shading marks masked frames.}
\label{fig:occlusion}
\end{figure}
\noindent\begin{minipage}[t]{0.48\textwidth}
\vspace{0pt}
\centering
\captionof{table}{\textbf{Six-clip truncation and occlusion.} Median WA-MPJPE (mm); body coverage is \name{} / Human3R. Shading compares methods within each condition.}
\label{tab:occlusion6}
\scriptsize
\setlength{\tabcolsep}{2.5pt}
\begin{tabular}{lccc}
\toprule
condition & \name{}$\downarrow$ & Human3R$\downarrow$ & coverage \\
\midrule
clean            & \bestc{63.6}  & \secondc{89.1}  & $100/100\%$ \\
legs cut         & \bestc{119.5} & \secondc{238.1} & $100/100\%$ \\
head cut         & \bestc{80.9}  & \secondc{131.8} & $48/52\%$ \\
moving occluder  & \bestc{71.3}  & \secondc{181.6} & $82/82\%$ \\
frame dropout    & \bestc{183.6} & \secondc{254.4} & $80/80\%$ \\
\bottomrule
\end{tabular}
\end{minipage}\hfill
\begin{minipage}[t]{0.49\textwidth}
\vspace{0pt}
\textbf{Six-sequence evaluation.} \currentrevision{Table}~\ref{tab:occlusion6} \currentrevision{reports the median over six EMDB-2 clips under identical synthetic masks.} \name{} \currentrevision{has lower error in 29 of 30 clip-condition pairs. Leg removal mainly worsens placement; head removal roughly halves detection coverage for both systems. Blank frames are hardest because camera and scale are shared across the window.}
\end{minipage}
\par\smallskip
\currentrevision{The crop-based mesh branch preserves local articulation under these masks; the larger changes are in person detection and placement. The masks are synthetic, so the table tests sensitivity to missing input rather than natural-occlusion frequency.}

\FloatBarrier
\needspace{0.52\textheight}
\section{Measured Failure Modes}
\label{app:limitations}
\begin{figure}[H]
\centering
\includegraphics[width=\textwidth]{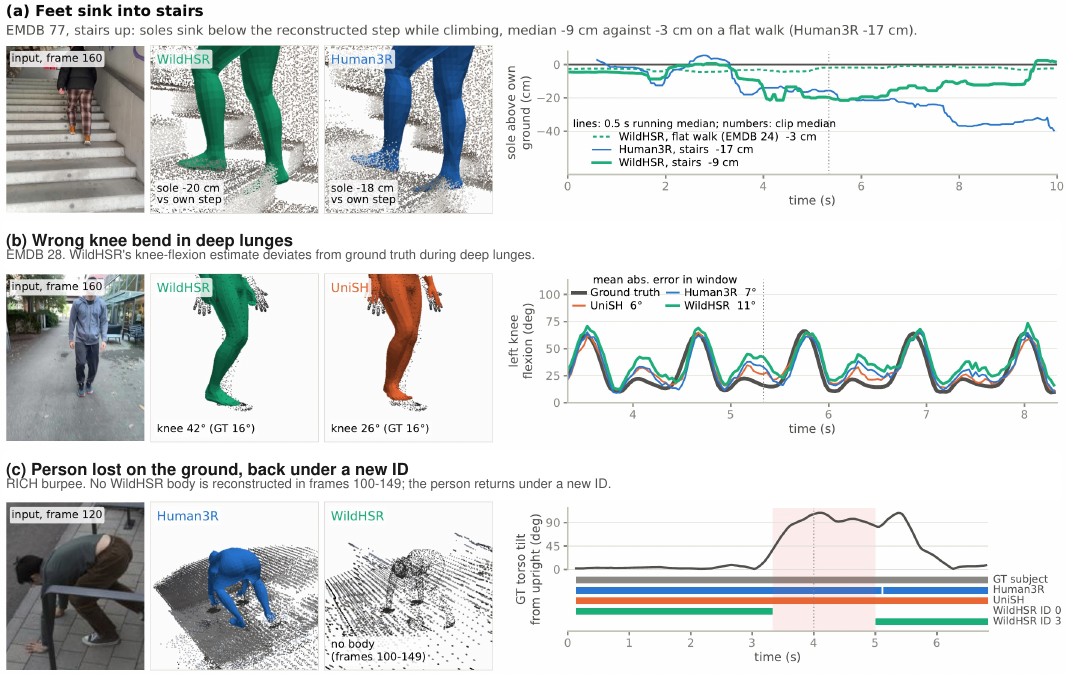}\vspace{5pt}
\caption{\textbf{Measured failure modes.} (a) Stair contact, (b) knee articulation, and (c) identity continuity during a burpee. Distances use each method's own metric scale.}
\label{fig:limitations}
\end{figure}
\textbf{What the failures show.} \currentrevision{(a) On EMDB 77, the soles sink below the reconstructed step while climbing: a median of} $-9$\,cm \currentrevision{over the clip and about} $-20$\,cm \currentrevision{on the steepest stretch, versus} $-3$\,cm \currentrevision{on a flat walk; Human3R also sinks} ($-17$\,cm)\currentrevision{. Scene contact is a training term, not an inference-time correction. (b) On EMDB 28, full-clip knee-flexion error is} $7.8^\circ$ \currentrevision{for} \name{} \currentrevision{versus} $5.4^\circ$ \currentrevision{for Human3R; the plotted window reports its own error. (c) In a RICH burpee, no WildHSR body is reconstructed for frames 100--149, and the person returns under a new track ID. These examples expose limits in stair placement, body articulation and track continuity.}

\FloatBarrier

\end{document}